\PassOptionsToPackage{table,dvipsnames}{xcolor}
\documentclass[10pt,twocolumn,letterpaper]{article}

\usepackage[pagenumbers]{iccv} % To force page numbers, e.g. for an arXiv version

\definecolor{iccvblue}{rgb}{0.21,0.49,0.74}
\usepackage[pagebackref,breaklinks,colorlinks,allcolors=iccvblue]{hyperref}
\usepackage{algorithm}
\usepackage{algorithmic}
\usepackage{multirow}
\usepackage[accsupp]{axessibility}  % Improves PDF readability for those with disabilities.

\def\paperID{11827} % *** Enter the Paper ID here
\def\confName{ICCV}
\def\confYear{2025}

\title{Task-Specific Zero-shot Quantization-Aware Training for Object Detection}

\author{
\begin{tabular}[t]{ccc}
\parbox[c]{0.3\textwidth}{\centering
Changhao Li\textsuperscript{1}\thanks{Equal contribution} \\
Atlanta, USA \\
{\tt\footnotesize cli911@gatech.edu}
} &
\parbox[c]{0.3\textwidth}{\centering
Xinrui Chen\textsuperscript{2}\footnotemark[1] \\
Shenzhen, China \\
{\tt\footnotesize cxr22@tsinghua.org.cn}
} &
\parbox[c]{0.3\textwidth}{\centering
Ji Wang\textsuperscript{3}\footnotemark[1] \\
Beijing, China \\
{\tt\footnotesize wangji20@tsinghua.org.cn}
} \\ [5ex]
\multicolumn{3}{c}{
\begin{tabular}{cc}
\parbox[c]{0.3\textwidth}{\centering
Kang Zhao\textsuperscript{4} \\
Beijing, China \\
{\tt\footnotesize zhaok14@tsinghua.org.cn}
} &
\parbox[c]{0.3\textwidth}{\centering
Jianfei Chen\textsuperscript{4}\thanks{Corresponding Author} \\
Beijing, China \\
{\tt\footnotesize jianfeic@tsinghua.edu.cn}
}
\end{tabular}
} \\ [5ex]
\end{tabular}
\\
\textsuperscript{1} School of Computational Science and Engineering, Georgia Institute of Technology \\
\textsuperscript{2} Shenzhen International Graduate School, Tsinghua University \\
\textsuperscript{3} School of Software, Tsinghua University \\
\textsuperscript{4} Dept. of Comp. Sci. and Tech., Institute for AI, Tsinghua-Bosch Joint ML Center, Tsinghua University \\
}

\begin{document}
\maketitle
\begin{abstract}

Quantization is a key technique to reduce network size and computational complexity by representing the network parameters with a lower precision. Traditional quantization methods rely on access to original training data, which is often restricted due to privacy concerns or security challenges. Zero-shot Quantization (ZSQ) addresses this by using synthetic data generated from pre-trained models, eliminating the need for real training data.
Recently, ZSQ has been extended to object detection. However, existing methods use unlabeled task-agnostic synthetic images that lack the specific information required for object detection, leading to suboptimal performance. In this paper, we propose a novel task-specific ZSQ framework for object detection networks, which consists of two main stages. First, we introduce a bounding box and category sampling strategy to synthesize a task-specific calibration set from the pre-trained network, reconstructing object locations, sizes, and category distributions without any prior knowledge. Second, we integrate task-specific training into the knowledge distillation process to restore the performance of quantized detection networks.
Extensive experiments conducted on the MS-COCO and Pascal VOC datasets demonstrate the efficiency and state-of-the-art performance of our method. Our code is publicly available at \href{https://github.com/DFQ-Dojo/dfq-toolkit}{https://github.com/DFQ-Dojo/dfq-toolkit}.

\end{abstract}    
\begin{figure*}[t]
% \vspace{-0.1cm}
    \centering
        \centerline{\includegraphics[width=\linewidth]{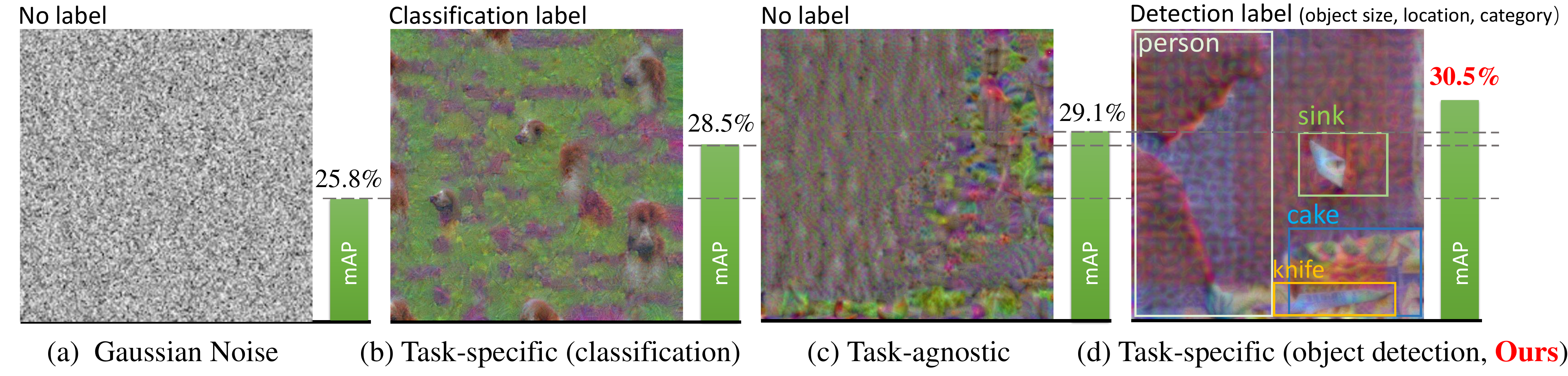}}
    \caption{Comparative analysis of different synthetic images and their impact on zero-shot quantization-aware training with object detection network Mask R-CNN. It is observed that task-specific calibration set matters and results in better performance on the MS-COCO dataset. Inspired by this, we integrate novel task-specific manners into data synthesis and quantized detection network finetuning processes.}
    \label{fig: mot}
    % \vspace{-0.5cm}
\end{figure*}

% \vspace{-0.5cm}
\section{Introduction}\label{sec: intro}

Object detection neural networks are integral to a wide array of computer vision applications, ranging from autonomous driving to surveillance systems~\citep{mao20233d, balasubramaniam2022object, oguine2022yolo, mishra2016study}. As the demand for deploying deep neural networks on resource-constrained devices continues to grow, quantization has emerged as a critical technique to reduce network size and computational complexity while maintaining performance~\citep{chen2019v2, deng2020model, han2015deep, wang2022makes}. However, traditional quantization-aware training methods often require access to the entire original training data, which may become inaccessible if the data are very huge or subjected to privacy protection.~\citep{krishnamoorthi2018quantizing,nagel2021white}. In these scenarios, Zero-shot Quantization (ZSQ)~\citep{cai2020zeroq, nagel2019data, yvinec2023spiq, xu2020generative, liu2021zero} presents a promising solution, enabling the quantization of neural networks without relying on original training data. These methods primarily train their model on synthetic data generated by inverting the network with randomly sampled labels.

Most prior work in this area has focused on classification networks. Specifically, they use model inversion to synthesize data by optimizing the gauss noise image through gradient descent. Their loss functions are often designated to classification tasks, which generate classification task-specific images. For example, GDFQ \citep{xu2020gdfq} introduced a knowledge-matching generator to synthesize label-oriented data using cross-entropy loss and batch normalization statistics (BNS) alignment. TexQ \citep{chen2024texq} emphasized the detailed texture feature distribution in real samples and devised texture calibration for labeled image generation. PSAQ-ViT \cite{li2022patch} introduced a patch similarity aware strategy to invert labeled images from Vision Transformers for quantization. These methods leverage synthetic data generated from the full-precision network to calibrate (\ie, post-training quantization) or finetune the quantized network (\ie, quantization-aware training) for accurate quantization. A more comprehensive discussion of data-driven quantization and ZSQ are presented in Appendix~\ref{sec: related}.

Recently, ZSQ has been extended to downstream tasks such as object detection, but its application is limited due to inherent complexity. Classification tasks require only a randomly assigned category as the target label, but object detection demands that the target label comprise both the bounding box location and the classification label, making it difficult to determine. Consequently, existing approaches for detection networks drop the detection training loss and instead adopt a task-agnostic strategy for data generation and model quantization. For instance, PSAQ-ViT V2~\citep{li2023psaq} introduced an adaptive teacher–student strategy to generate task-agnostic images for finetuning the quantized model via knowledge distillation. Similarly, MimiQ~\citep{choi2024mimiq} proposed inter-head attention similarity and applied head-wise structural attention distillation to align the attention maps of the quantized network with those of the full-precision teacher across downstream tasks. CLAMP-ViT~\citep{ramachandran2024clamp} employed a two-stage approach, cyclically adapting between data generation and quantization. Although task-agnostic strategy enhances generalizability across different downstream tasks, it leads to a lack of task-specific bounding box size and location information in the object detection network, potentially resulting in suboptimal performance.

We argue that incorporating task-specific information into ZSQ can significantly increase its effect. By augmenting training loss with object categories and bounding box information, our method can outperform previous task-agnostic methods and, in some settings, may even achieve comparable results to networks trained with full real-data.

% ignore the training losses during image generation, resulting in task-agnostic unlabeled images. With the unlabeled synthetic images, these methods

The proposed task-specific framework consists of two stages. In the generation stage, we introduce a novel bounding box and category sampling strategy to synthesize a calibration set from a pre-trained detection network, which reconstructs the location, size and category distribution of objects within the data without any prior knowledge. In the quantization stage, we integrate the detection training loss into the distillation process to further amplify the efficacy of quantized detection network finetuning.

Extensive experiments on MS-COCO and Pascal VOC confirm the state-of-the-art performance of our method. For example, when quantizing YOLOv5-l to 6-bit, we achieve a 1.7\% mAP improvement over LSQ trained with full real data. Furthermore, tests on YOLO11 and Swin Transformer models show our approach surpasses task-agnostic ZSQ  by 2-3\% in mAP across various quantization settings. Specifically, our contributions are threefold:

\begin{enumerate}
\item   \textbf{Drawback of task-agnostic calibration is revealed.} 
% We demonstrate that task-agnostic or task-mismatching images are detrimental to quantization. Despite the demonstrated efficiency of task-agnostic synthetic images for ZSQ across downstream tasks, they overlook significant task-specific information, which we find to be crucial for quantized model performance.

We emphasize task-specific synthetic images for zero-shot quantization of object detection networks. By developing a task-specific approach that optimizes both data synthesis and finetuning, we unlock the full performance potential of quantized object detection networks.

\item  \textbf{Task-specific object detection images synthesis.} 
We propose a bounding box sampling method tailored for object detection networks to reconstruct object categories, locations, and sizes in synthetic samples without any prior knowledge.
% To inverse the category-imbalance object detection model, we employ a relabel strategy for category sampling strategy in the calibration set generation process and reconstruct the category distribution of objects within the synthetic samples without any prior knowledge. 

\item  \textbf{Task-specific quantized network distillation.}
We integrate object detection task-specific finetuning into quantized network distillation, effectively restoring the performance of quantized object detection networks.

\end{enumerate}

\section{Motivation}\label{sec: motiv}

\subsection{Preliminary on Quantization}

Network quantization has emerged as a critical technique for reducing network size and computational cost while maintaining performance~\citep{chen2019v2, deng2020model, han2015deep, wang2022makes}.

Given a floating-point tensor $w_{fp}$~(weights or activations) and quantization bit width $b$, a commonly used per-tensor symmetric quantizer LSQ~\citep{esser2019learned} for both weights and activations to quantize the data $\hat{w_{fp}}$ can be defined as:

% \vspace{-1em}
%
\begin{equation}\label{eq:quantization1}
w_{int} = clip(\lfloor \frac{w_{fp}}{s} , -2^{b-1}, 2^{b-1}-1 \rceil),	
\end{equation}

\vspace{-1em}
\begin{equation}
	\label{eq:quantization2}
	\hat{w_{fp}} = w_{int} \times s.
\end{equation}

Here, $w_{int}$ denotes the quantized integer representation of the data, $\lfloor input \rceil$ rounds the input to its nearest integer, and the step size $s$ is a quantization parameter that is obtained with calibration set and updated during quantization-aware training. Notably, the calculation of quantization parameters requires access to real training data as the calibration set, rendering traditional quantization methods inapplicable when training data is unavailable.

% However, the algorithms designed for classification tasks may not be directly applicable to detection tasks, as they cannot effectively utilize the output of the detection head. From the perspective of synthetic samples with detection networks, ~\cite{chawla2021data,chengeodiffusion,Wang_2024_CVPR} presented relevant methods to synthesize data with ground-truth or small amounts of real data for distillation or network training. However, both ground-truth and real samples should be prohibited in the zero-shot settings and they were not aimed at zero-shot quantization tasks and therefore lacked consideration of the internal information of the model.

\subsection{Revisiting ZSQs for Object Detection} \label{subsec: revisit zsq}
% \paragraph{Task-agnotic calibration images.}
\paragraph{Task-specific calibration set matters.}

ZSQs synthesize images as the calibration set for quantization. Despite the proven efficacy of task-agnostic images in enhancing various downstream tasks within the ZSQ framework, it is intuitively evident that these images lose a considerable amount of task-specific information, which potentially compromise the performance of specific tasks.
As illustrated in Fig. \ref{fig: mot}, we visualize four types of synthetic images utilized for zero-shot quantization-aware training: Gaussian noise, classification task-specific, task-agnostic, and object detection task-specific images. The Gaussian noise image serves as a baseline for comparison. Task-agnostic images capture the general features extracted by the network's backbone, whereas task-specific images are derived through model inversion with corresponding labels, such as classification categories and object detection bounding box localizations.
The visualization analysis in Fig.~\ref{fig: mot} reveals that task-specific images extract a richer set of features compared to other types of images, including object location, label, and size. Subsequently, we conduct a comparative analysis of synthetic images to explore their efficacy in zero-shot quantization-aware training, indicating that task-mismatched images lead to performance degradation, as demonstrated by the classification task-specific image in Fig.~\ref{fig: mot}(b), whereas task-specific images enhance performance, as shown by the object detection task-specific image in Fig.~\ref{fig: mot}(d). Therefore, task-specific calibration is crucial in the zero-shot quantization for detection tasks.

\vspace{-1em}
\paragraph{Challenges on task-specific ZSQ for detection.}

% While task-specific zero-shot quantization has achieved remarkable success in classification tasks, extending it to object detection faces unique challenges due to the inherent complexity of object detection, which encompasses both localization and classification subtasks.
While task-specific zero-shot quantization has achieved remarkable success in classification tasks, extending it to object detection faces significant challenges.

First, the label sampling methods in the classification task for synthesizing labeled images cannot be directly extended to object detection. For classification networks, data synthesis typically requires only a randomly sampled category ID label~\citep{choi2021qimera, zhong2022intraq, qian2023adaptive, qian2023rethinking, chen2024texq}. In contrast, for object detection, the location and size of objects within the samples remain unknown and elusive in zero-shot scenarios, making artificial reconstruction without ground-truth information extremely challenging.
Besides, object detection datasets typically exhibit significant category imbalance, which is not captured by random category sampling methods designed for classification tasks (e.g., category-balanced ImageNet or CIFAR-10/CIFAR-100). Therefore, randomly sampling object categories, locations and sizes often results in implausible category distribution, relative positions and sizes, leading to unrealistic synthetic data.
Furthermore, the task-specific finetuning strategy for detection networks using synthetic calibration data remains underexplored. The currently used logits alignment methods, which are designed for classification networks, may not be sufficient for the more complex object detection networks. This insufficiency makes it difficult to fully leverage the limited synthetic data efficiently, thereby hindering the performance improvement of quantized detection models.
\section{Methodology}\label{sec: method}

\begin{figure*}[t]
% \vspace{-0.1cm}
    \centering
        \centerline{\includegraphics[width=0.99\linewidth]{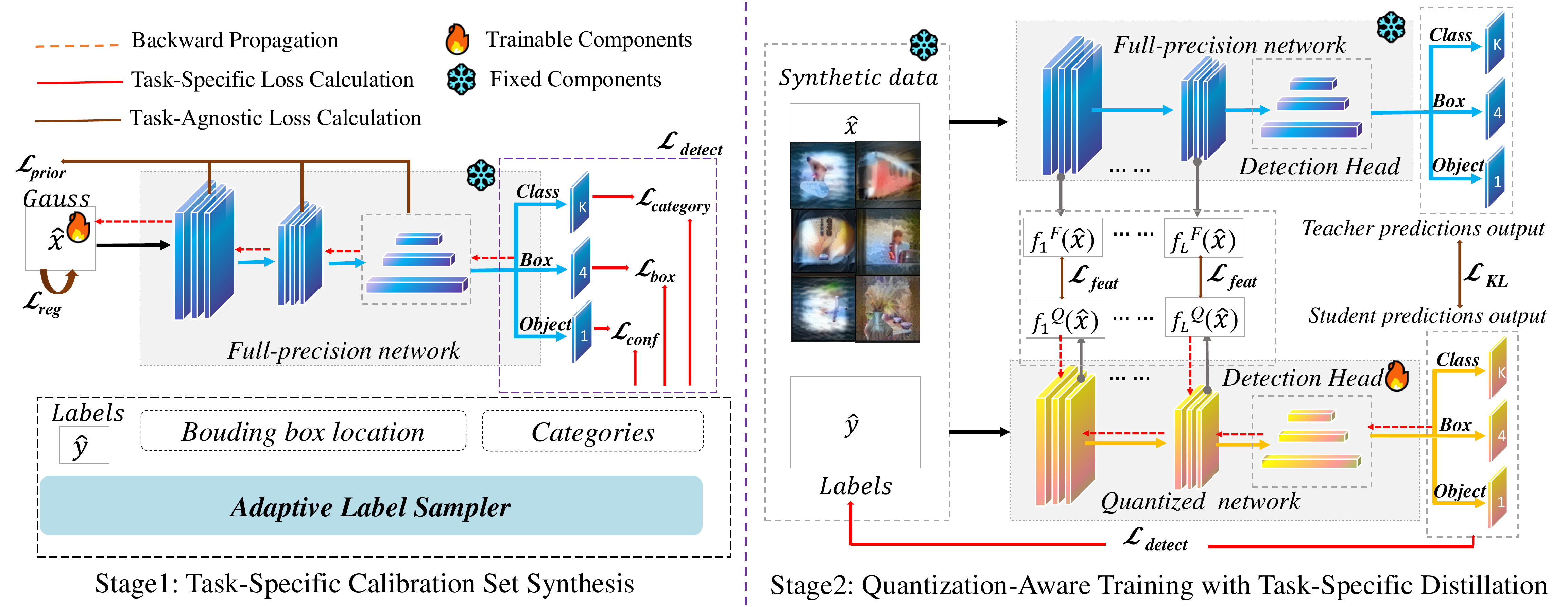}}
    \caption{Overall architecture of our method. Our framework comprises: 1) constructing a task-specific condensed calibration set and 2) conducting quantization-aware training with task-specific distillation. See Section~\ref{sec: method} for details.}
    \label{fig: framework}
    \vspace{-0.5cm}
\end{figure*}

In this section, we provide an overview of the proposed framework in Fig.~\ref{fig: framework}. It contains two stages: generating a task-specific calibration set and performing quantization-aware training (QAT) with task-specific distillation.

\subsection{Preliminaries on Task-Agnostic Data Synthesis}\label{subsec: task-agnostic data synthesis}
The calibration set utilized in model quantization needs to reflect the inherent distribution of model. Zero-shot quantization seeks to generate a synthetic calibration set that matches the model's distribution~\citep{cai2020zeroq}.
The synthetic calibration set can be derived through noise optimization~\citep{cai2020zeroq, zhong2022intraq, zhang2021diversifying}, which is usually instantiated by distribution approximation~\citep{cai2020zeroq, xu2020generative}. Existing zero-shot methods in object detection typically require generating a calibration set the same size as the training set (120k images for MS-COCO)~\citep{chawla2021data}. In contrast, we only generate a small amount of calibration set to extract features of the data.

Given a batch of $N$ inputs $x \in R^{N \times3\times H \times W}$, where each pixel is initialized from random Gaussian noise $\text{x}_{i,c,h,w} \sim \mathcal{N}(0,1)$, and a pre-trained full-precision detection network $\phi(\theta)$, synthetic calibration set are obtained through optimizing the inputs to match the batch normalization statistics~(BNS)~\citep{yin2020dreaming} in convolutional neural networks (\eg, YOLO~\citep{redmon2016you}, Mask R-CNN~\citep{he2017mask}):

\vspace{-1.5em}
\begin{equation}
   \min_{x} \mathcal{L}_{prior}(x) 
    = \sum_{l=1}^L(|| \mu^{l}(\theta, x) - \mu^{l}(\theta)||_2   + ||{\sigma^l}(\theta, x) - {\sigma^{l}(\theta)||_2}), \small
\label{eqn:bn loss}
\end{equation}

where $ \mu^{l}(\theta) / \sigma^{l}(\theta)$ are mean/variance parameters stored in the $l$-th BN layer of $\phi(\theta)$ and $ \mu^{l}(\theta, x) / \sigma^{l}(\theta, x)$ are mean/variance statistic calculated on inputs using $\phi(\theta)$. It enforces feature similarities at all levels by minimizing the distance between the feature map statistics for the synthesized image $x$ and the real image.  For vision transformer models (\eg, ViT~\citep{dosovitskiy2020image}, Swin Transformer~\citep{liu2021swin}), since they only use layer normalization (LN) and do not store any runtime statistical information, we adopt the Patch Similarity Entropy loss~($L_{PSE}$), as described in~\citep{li2022patch}, as $L_{prior}$ to align inputs with the original data.

Besides the statistical alignment objective function, a regularization term consisting of the total variance and $l_2$ norm of the input image is always involved in the final loss function to steer images away from unrealistic images~\citep{mahendran2015understanding}:

\begin{equation}
\min_x\mathcal{L}_{reg}(x) = \alpha_{TV} \mathcal{L}_{TV}(x) + \alpha_{l_2} \| x \|_2^2,
\label{eqn:prior loss}
\end{equation}

\noindent where $\mathcal{L}_{TV}$ promotes similarity between adjacent pixels by minimizing their Frobenius norm, consequently enhancing the smoothness, $\alpha_{TV}$ and $\alpha_{l_2}$ are hyper-parameters balancing the importance of two terms. Finally, we can regard task-agnostic data synthesis as a regularized minimization problem and optimize the following function:

\begin{equation}
\min_{x} ~\alpha_{prior} \mathcal{L}_{prior}(x) +  \mathcal{L}_{reg}(x).
\label{eqn:main loss pre}
\end{equation}

\subsection{Stage I: Task-Specific Calibration Set Synthesis}\label{subsec: task-specific data synthesis}
\begin{figure}[t]
% \vspace{-0.1cm}
    \centering
        \centerline{\includegraphics[width=1\linewidth]{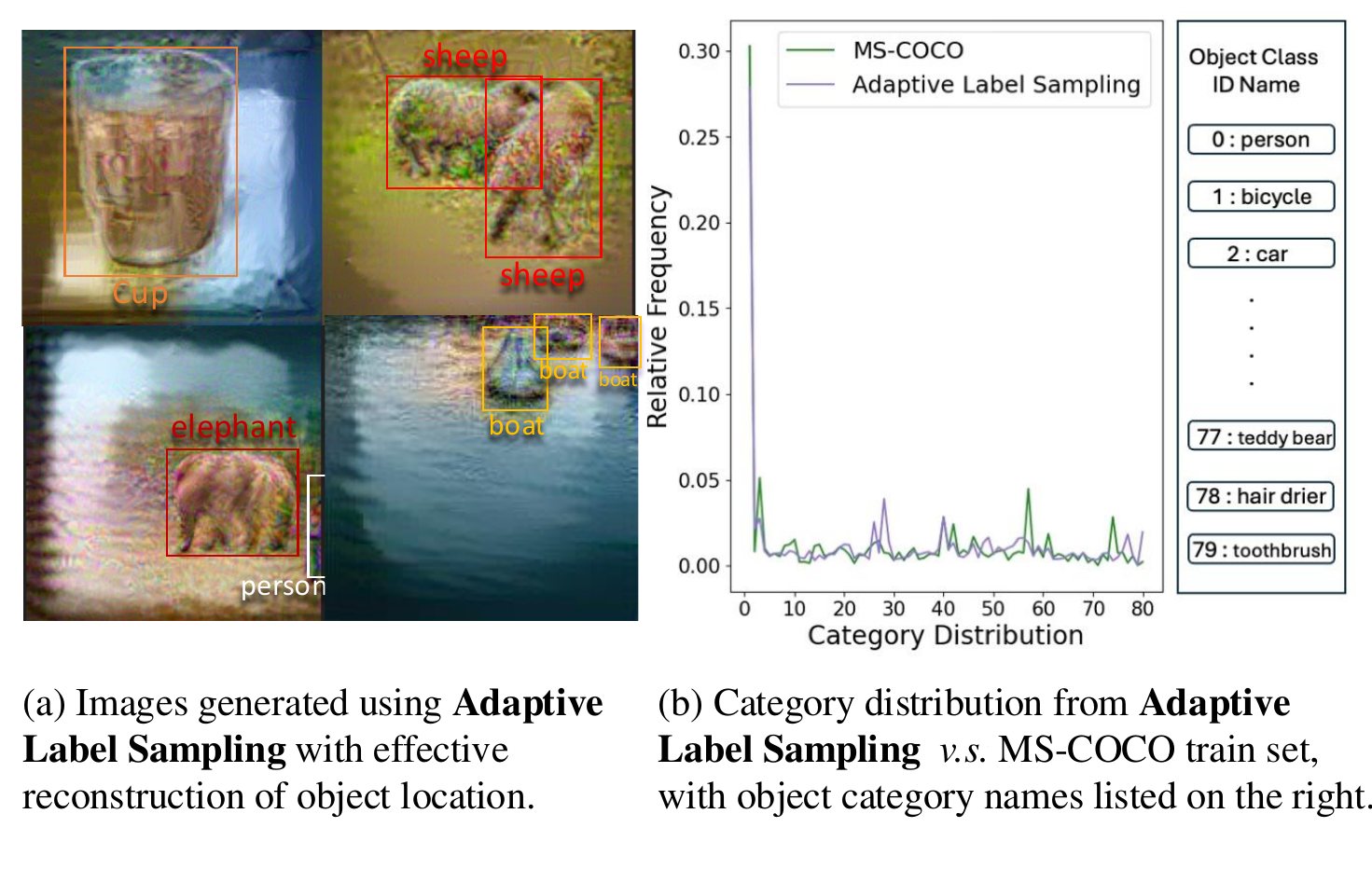}}
    \caption{(a) Images generated by \textbf{Adaptive Label Sampling} on a YOLOv5 detector pre-trained on MS-COCO. (b) \textbf{Adaptive Label Sampling} can generate a category distribution frequency similar to MS-COCO in a zero-shot setting.}
    \label{fig: cons}
    \vspace{-0.5cm}
\end{figure}
% \vspace{-3em}
\paragraph{Task-Specific Loss Calculation}
As discussed in Section~\ref{subsec: revisit zsq}, the object detection task requires labels that include the precise location and size of objects within an image, making artificial reconstruction extremely challenging. Consequently, previous zero-shot quantization approaches for object detection commonly utilize a task-agnostic loss, as described in Eq.~\ref{eqn:main loss pre}, to generate images~\citep{choi2024mimiq, ramachandran2024clamp, li2023psaq}. However, our experiments indicate that this approach leads to a significant loss of task-specific information, ultimately resulting in degraded performance on downstream tasks. Therefore, in this section, we introduce the training loss of the object detection network to optimize the sampled inputs, aiming to recover task-specific information.

Given a full-precision detection network $\phi(\theta)$ and a Gaussian-initialized input $x$, the standard form of the training loss in classification networks is expressed as $L_{classify}(\phi(x), c)$, where the target label $c$ is an integer generated through random sampling. However, in object detection tasks, label information is more complex, often encompassing the object's position and size. This can be formulated as $L_{detect}(\phi(x), y)$, which typically consists of three components: a box category loss $\mathcal{L}_{category}$, a box dimension loss $\mathcal{L}_{box}$, and a grid location loss $\mathcal{L}_{conf}$. Through this formulation, we can reconstruct the category and coordinate information of bounding boxes in real images. Combining with the task-agnostic loss, we derive the following task-specific data synthesis function:

\vspace{-1em}
\begin{equation}
\min_{x} ~\alpha_{prior} \mathcal{L}_{prior}(x) +  \alpha_{detect} \mathcal{L}_{detect}(\phi(x), \text{y}) +  \mathcal{L}_{reg}(x).
\label{eqn:main loss}
\end{equation}
\vspace{-2em}

\paragraph{Adaptive Label Sampling}

To obtain task-specific information in a zero-shot scenario, we propose an adaptive label sampling method for label synthesis that effectively extracts essential bounding box coordinates and category information for object detection. Our approach relies solely on a pre-trained network and does not require any additional information (\eg, metadata, feature activations) or extra networks (\eg, pre-trained generative models).

Motivated by \citep{yin2024squeeze}, which integrates soft labels into the data reconstruction process to better align synthetic data and labels, we propose an adaptive label sampling method, detailed as follows:

We start by randomly generating a label containing a single object, where the bounding box coordinates and category are uniformly sampled within a valid range~(Details in Table~\ref{table: One box sampling}). This label is then used as the ground truth target, and the input is optimized using Eq.~\ref{eqn:main loss}. After a fixed number of iterations, we re-detect objects in the input using a pre-trained full-precision detection network. High-confidence regions from the teacher network’s feature map are added as new labels, while low-confidence regions are removed, ensuring that each image retains at least one label. The overall process is detailed in Algorithm~\ref{alg:adaptive sampling}.

The input image and target labels are updated alternately, progressively aligning with each other throughout the process. We also provide a visualization of this alignment in Fig.~\ref{fig: relabel framework}.
With this sampling strategy, we eliminate the need for real detection labels. Besides, as presented in Fig.~\ref{fig: cons}(b), the approach can eventually produce bounding box categories that closely resemble the actual distribution, while also reconstructing objects' relative positions, sizes, and counts. This capability enables tackling the challenging zero-shot quantization-aware training for object detection.

\paragraph{Task-Specific Data Synthesis}
Through adaptive label sampling, we obtain labels that resemble the bounding box category and coordinate information in real images. We then fix the generated labels and optimize a newly Gaussian-initialized input toward these targets using the task-specific training loss introduced in Eq.~\ref{eqn:main loss}. See Appendix~\ref{subsec: adaptive label sampling} for more details.

Compared to randomly sampling multiple bounding box coordinates and category labels, our adaptive label sampling method effectively utilizes the knowledge embedded in pre-trained object detection networks, resulting in higher-quality image synthesis. As illustrated in Fig.~\ref{fig: cons}(a), images generated using our approach accurately capture object locations within the image, producing more realistic results. Further studies, as discussed in Appendix~\ref{sec: additional qualitative results}, demonstrate that our method yields clearer objects and more coherent layouts than those generated through random sampling.

\begin{table*}[t]

\caption{Comparison with real data QATs on YOLOv5/YOLO11 on MS-COCO validation set.}
\label{table: Comparison on YOLO}
\centering

\resizebox{\linewidth}{!}{
        \begin{tabular}{llllcccccc}
        \toprule
        ~ & ~ & ~ & ~ & \multicolumn{6}{c}{mAP / mAP50} \\
        \cmidrule(r){5-10}
        Method & Real Data & Num Data & Prec. & YOLOv5-s & YOLOv5-m & YOLOv5-l & YOLO11-s & YOLO11-m & YOLO11-l \\
        \midrule
        Pre-trained & \checkmark & 120k(full) & FP & 37.4/56.8 & 45.4/64.1 & 49.0/67.3 & 47.0/65.0 & 51.5/70.0 & 53.4/72.5 \\
        \midrule
         LSQ & \checkmark & 120k(full) & \multirow{5}{*}{W8A8} & 35.7/54.9 & 43.2/62.2 & 46.0/64.9 & 44.9/61.8 & 49.1/66.2 & 50.4/67.4\\
         LSQ+ & \checkmark & 120k(full) &  & 35.4/54.6 & 43.3/62.4 & 46.3/64.9 & 45.1/61.8 & 49.6/66.7 & 50.9/67.7 \\
         LSQ & \checkmark & 2k &  & 31.6/50.6 & 36.5/55.6 & 40.3/59.1 & 44.0/60.8 & 47.6/64.5 & 48.8/65.8 \\
         LSQ+ & \checkmark & 2k &  & 31.5/50.3 & 36.6/55.8 & 40.1/58.6 & 43.8/60.7 & 
 47.8/64.7 & 48.5/65.3 \\
         \rowcolor{teal!10} \textbf{Ours} & \texttimes & 2k &  & \textbf{35.8}/\textbf{55.0} & \textbf{43.6}/\textbf{62.3} & \textbf{47.3}/\textbf{65.6} & \textbf{45.6/62.3} & \textbf{50.0/66.5} & \textbf{51.8/68.4}\\
        \midrule
         LSQ & \checkmark & 120k(full) & \multirow{5}{*}{W6A6} & 31.5/49.9 & 41.3/60.0 & 43.3/62.1 & 43.0/59.7 & 47.4/64.2 & 48.6/65.3\\
         LSQ+ & \checkmark & 120k(full) &  & 32.3/50.9 & \textbf{41.3}/\textbf{60.3} & 43.4/62.3 & \textbf{43.2/59.8} & \textbf{47.6/64.3} & \textbf{48.9/65.8} \\
         LSQ & \checkmark & 2k &  & 28.9/47.2 & 35.0/53.9 & 37.7/55.7 & 41.5/58.3 & 45.0/61.9 & 45.8/62.5 \\
         LSQ+ & \checkmark & 2k &  & 28.6/46.7 & 34.2/52.6 & 37.5/55.8 & 41.6/58.2 & 44.8/61.7 & 45.9/62.8 \\
         \rowcolor{teal!10} \textbf{Ours} & \texttimes & 2k &  & \textbf{32.7}/\textbf{51.4} & 41.0/59.7 & \textbf{45.1}/\textbf{63.3} & 43.0/59.3 & 47.1/63.2 & 48.4/64.6 \\
         \midrule
         LSQ & \checkmark & 120k(full) & \multirow{5}{*}{W4A8} & 32.2/51.0 & 41.0/59.9 & 44.6/63.5 & 42.4/59.1 & 47.6/64.4 & 48.7/65.6 \\
         LSQ+ & \checkmark & 120k(full) &  & 32.3/51.1 & 41.2/60.1 & 44.4/63.2  & \textbf{42.7/59.3} & \textbf{47.8/64.8} & 49.4/66.3 \\
         LSQ & \checkmark & 2k & & 28.1/46.5 & 35.8/54.6 & 39.0/57.5 & 40.9/57.5 & 45.2/62.4 & 46.1/63.0 \\
         LSQ+ & \checkmark & 2k & & 29.3/47.8 & 37.8/56.9 & 40.6/59.7 & 40.7/57.3 & 45.2/62.3 & 46.4/63.4 \\
         \rowcolor{teal!10} \textbf{Ours} & \texttimes & 2k &  & \textbf{33.0}/\textbf{52.5} & \textbf{42.6}/\textbf{61.7} & \textbf{46.2}/\textbf{64.7} & 42.6/58.9 & 47.7/64.1 & \textbf{49.4/65.7} \\
         \bottomrule
        \end{tabular}
}

\end{table*}

\subsection{Stage II: QAT with Task-Specific Distillation} \label{subsec: Efficient fine-tuning with distillation}

In this section, we propose to minimize the knowledge gap between the full-precision pre-trained network (teacher) and the low-precision quantized network (student) through knowledge distillation.

Knowledge distillation~\citep{hinton2015distilling} is a widely used technique for knowledge transfer. Previous studies~\citep{ding2023cbq, li2023hard} have used it to enhance performance in classification tasks involving quantized CNNs and LLMs. However, the hidden states from backbone and prediction head of a full-precision pre-trained YOLO network contain much of the statistical information from real training data~\citep{yin2020dreaming}, which is challenging to fully utilize. To address this, we propose using feature-level distillation to align intermediate features and prediction-matching distillation to ensure consistency between the predictions of the quantized and pre-trained networks.

\paragraph{Prediction-matching Distillation} As proposed in Section~\ref{subsec: task-specific data synthesis}, our synthetic calibration set $\{(\hat{x}^i, \hat{y}^i)\}_{i=1}^N$ is the result of the network backpropagating through pre-defined labels, directly aligning predictions of the quantized network with the targets would lead to severe over-fitting issues. Therefore, we introduce the Kullback–Leibler (KL) divergence loss~\citep{kullback1951information} between the predictions of the quantized network and the full-precision network as soft labels to align their outputs, thereby recovering the performance of the quantized network, which is formulated as:

\vspace{-1.5em}
\begin{equation}
\begin{aligned}
\min_{\theta'} \mathcal{L}_{KD} &= \frac{\tau^2}{N} \sum_{i=1}^{N} KL(z^F(\hat{x}_i;\theta), z^Q(\hat{x}_i;\theta')),
\end{aligned}
\label{eqn:KL loss}
\end{equation}

\noindent where $\{\hat{x_i}\}_{i=1}^{N}$ is a batch of the calibration set images, $z^F(\hat{x}_i;\theta) /  z^Q(\hat{x}_i;\theta')$ are output predictions from full-precision / quantized network and $\tau$ is the distilling temperature. We denote parameters of full-precision / quantized network as $\theta / \theta'$.
\paragraph{Feature-level Distillation} 
We extend the knowledge transfer approach to the feature level by introducing a feature distillation method that explicitly aligns intermediate features between the teacher and student. This significantly improves training stability in low-bit settings, where QAT at ultra-low bit widths often leads to rapid error accumulation. By ensuring feature consistency between the teacher and student through feature distillation, we effectively reduce error propagation throughout the training process.

In the quantization-aware training stage, given a batch of synthetic image $\{\hat{x}_i\}_{i=1}^N$, we impose the mean squared error constraints between the feature maps from teachers and students. With $L$ being the number of distilling network layers, the feature distillation loss $\mathcal{L}_{feat}$ is expressed as:

\vspace{-1em}
\begin{equation}
\min_{\theta'} \mathcal{L}_{feat} =  \frac{1}{NL} \sum_{i=1}^{N} \sum_{l=1}^{L} || f_l^F(\hat{x}_i;\theta)- f_l^Q(\hat{x}_i;\theta')||_2^2.
\label{eqn:feature loss}
\end{equation}  
\vspace{-2em}

\paragraph{Task-Specific Quantization-Aware Training}
Previous works do not explicitly incorporate task-specific loss into their QAT training objectives~\citep{choi2024mimiq, ramachandran2024clamp, li2023psaq}. While this allows their networks to remain flexible and adaptable to various downstream tasks (\eg instance segmentation, object classification), we find that it often compromises performance on the target task — object detection in our case. To mitigate this, we introduce the task-specific training loss $L_{detect}$ during the QAT phase, enabling the quantized network to learn bounding box information directly from labels. %We observe this approach can significantly improve performance on the target task.

To this end, the total loss for quantization-aware training can be summarized as:

\vspace{-1em}
\begin{equation}
\min_{\theta'} \mathcal{L}^Q = \beta_{KL}  \mathcal{L}_{KD} +  \beta_{feat} \mathcal{L}_{feat} + \beta_{detect} \mathcal{L}_{detect},
\label{eqn:QAT loss}
\end{equation}

\noindent where $\beta_{KL}$,  $\beta_{feat}$ and $\beta_{detect}$ are hyper-parameters to balance the three terms.

\section{Experiments and Results}\label{sec: experiments}

\begin{figure}[t]
% \vspace{-0.1cm}
    \centering
        \centerline{\includegraphics[width=1\linewidth]{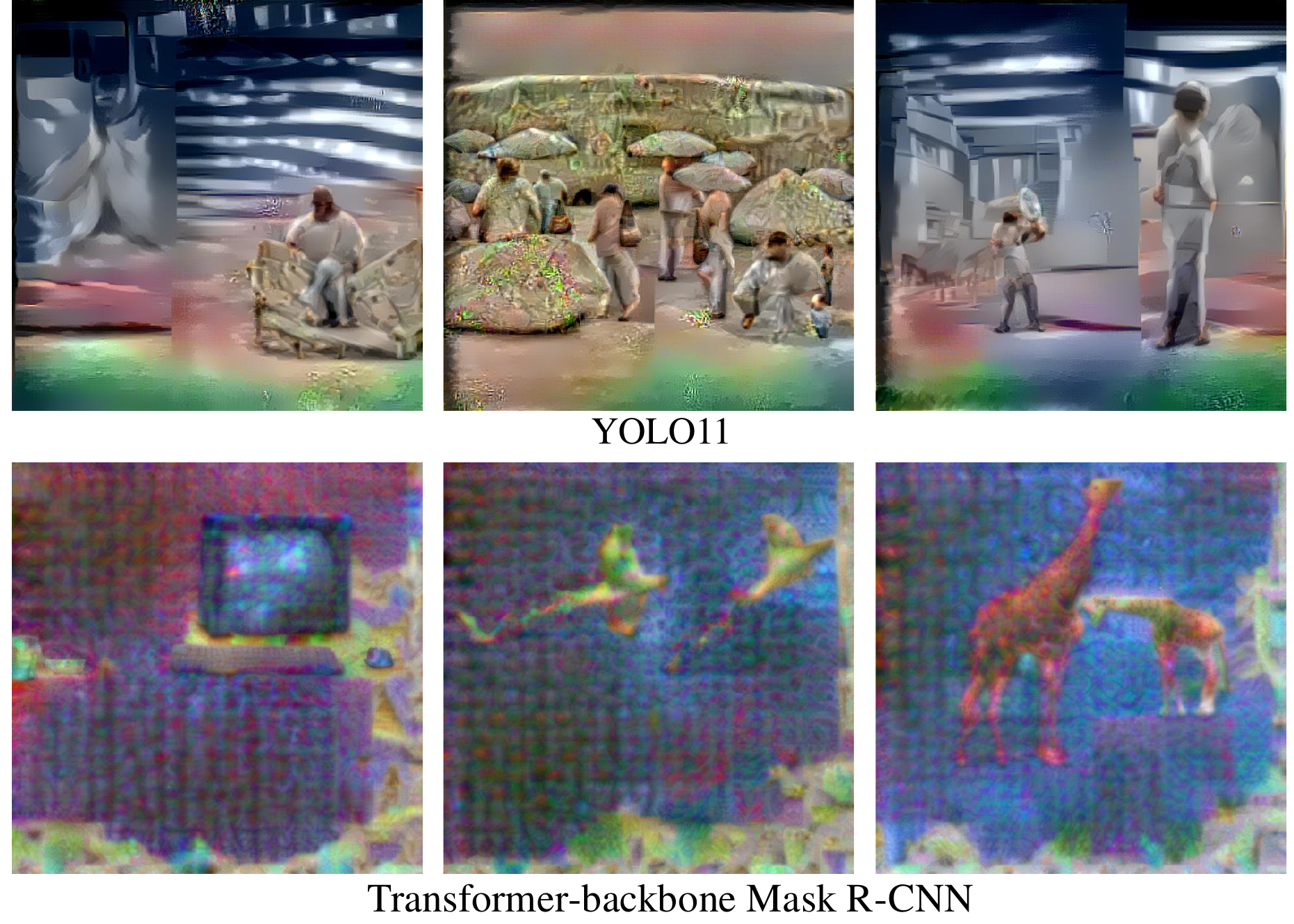}}
    \caption{Visualization of images generated by YOLO11 and Transformer-backbone Mask R-CNN. More examples can be found in Appendix~\ref{sec: additional qualitative results}.}
    \label{fig: img visualization_main}
    % \vspace{-0.5cm}
\end{figure}

In this section, we validate the proposed task-specific scheme on MS-COCO 2017~\citep{lin2014microsoft} 
 and Pascal VOC~\citep{everingham2010pascal} datasets. Following LSQ~\citep{esser2019learned}, we apply symmetric quantization to both weights and activations. Through extensive experiments, we demonstrate that our method is effective on various architectures, including the YOLOv5~\citep{ultralytics2021yolov5} series, YOLO11~\citep{Jocher_Ultralytics_YOLO_2023} series, CNN-based Mask R-CNN~\citep{he2017mask}, as well as Transformer-based Mask R-CNN~\citep{liu2021swin} for object detection tasks. We further compare our approach with standard quantization-aware training baselines, including LSQ~\citep{esser2019learned} and LSQ+~\citep{bhalgat2020lsq+}, using real data. Implementation details are provided in Appendix~\ref{sec: Implementation Details}, and ablation studies on different settings and components are presented in Appendix~\ref{sec: Ablation study}. We also present visualizations of images generated by various models, including YOLO11 and Transformer-based Mask R-CNN, as shown in Fig.~\ref{fig: img visualization_main}. More images and further analysis are provided in Appendix~\ref{sec: additional qualitative results}.

\subsection{Comparison with YOLO Networks} \label{subsubsec: Comparison with one-stage}

For YOLO object detection networks, we conduct experiments using the widely adopted YOLOv5-m/s/l models as well as the latest YOLO11-s/m/l. As baselines, we include competitive real-data QAT methods such as LSQ~\citep{esser2019learned} and LSQ+~\citep{bhalgat2020lsq+}. Extensive experiments show that our method outperforms both LSQ and LSQ+, which rely on full real-image training, by using only a small amount of ground truth label information. We compare our performance against LSQ and LSQ+, both trained on 120k real images from the MS-COCO dataset. In contrast, our approach utilizes only 2k ground truth labels for calibration set generation. The results are summarized in Table~\ref{table: Comparison on YOLO}, with further details on the performance of the YOLOv5 series models at lower precision provided in Appendix~\ref{full comparison with yolov5}.

\textbf{Bit-width } Our method demonstrates strong performance across different bit-widths. Notably, when quantizing both weights and activation parameters to 8-bit precision, we find that our approach outperforms full-data LSQ/LSQ+ across all YOLOv5 and YOLO11 models by 0.3\%–1.0\%. Even when the quantization precision is further reduced to 6-bit or lower, our method still achieves results comparable to full-data LSQ/LSQ+, while significantly surpassing LSQ/LSQ+ with the same data amount by 2\%–6\%.

\textbf{Network Size } Larger networks tend to exhibit poor performance with existing quantization-aware training methods, particularly in low-bit-width cases. For instance, in the 6-bit case, LSQ+ applies to YOLOv5-s resulting in a 5.1\% decrease in mAP compared to the pre-trained network, which achieves 5.6\% with YOLOv5-l. In contrast, our approach yields only a 4.7\% gap in mAP when quantizing YOLOv5-s to 6-bit precision, and the difference further reduces to 3.9\% with YOLOv5-l.  

\textbf{Efficiency } Our method performs quantization-aware training using a condensed synthetic detection calibration set, just 1/60 the size of the original, yet achieves superior performance compared to traditional QAT methods that require the full training dataset. This enables more efficient training, reducing computational costs and time while producing higher-quality low-precision models.

\subsection{Comparison with Mask R-CNN Networks}
\label{subsubsec: Comparison with two-stage}
\begin{table}[t]
 \caption{Comparison with real data QATs on CNN-based Mask R-CNN.}
\label{table: Comparison with Mask R-CNN}
\centering
% \begin{sc}
\resizebox{1\linewidth}{!}{
        \begin{tabular}{llllll}
        \toprule
        Dataset & Method & Real Data & Num Data & Precision & mAP\\
        \midrule
        \multirow{4}{*}{VOC} & Pre-trained & \checkmark & 5k(full) & FP32 & 75.6 \\
        \cmidrule(r){2-6}
         & LSQ & \checkmark & 5k(full) & \multirow{3}{*}{W8A8} & 72.4 \\
         % & LSQ+ & \checkmark & 120k(full) &  & 35.4 & 54.6 \\
         & LSQ & \checkmark & 50 &  & 70.9 \\
         % & LSQ+ & \checkmark & 2k &  & 31.5 & 50.3 \\
           &  \cellcolor{teal!10}\textbf{Ours} &  \cellcolor{teal!10}\texttimes &  \cellcolor{teal!10}50 &  \cellcolor{teal!10} &  \cellcolor{teal!10}\textbf{72.9} \\
         \midrule
         \multirow{7}{*}{MS-COCO} & Pre-trained & \checkmark & 120k(full) & FP32 & 38.1 \\
        \cmidrule(r){2-6}
         & LSQ & \checkmark & 120k(full) & \multirow{3}{*}{W8A8} & 35.0 \\
         & LSQ & \checkmark & 2k &  & 32.9 \\
        &  \cellcolor{teal!10}\textbf{Ours} &  \cellcolor{teal!10}\texttimes &  \cellcolor{teal!10}2k &  \cellcolor{teal!10} &  \cellcolor{teal!10}\textbf{35.2} \\
         \cmidrule(r){2-6}
         & LSQ & \checkmark & 120k(full) & \multirow{3}{*}{W4A8} & 34.6 \\
         & LSQ & \checkmark & 2k &  & 32.3 \\
          &  \cellcolor{teal!10}\textbf{Ours} &  \cellcolor{teal!10}\texttimes &  \cellcolor{teal!10}2k &  \cellcolor{teal!10} &  \cellcolor{teal!10}\textbf{34.6} \\
         \bottomrule
        \end{tabular}
        }
        % \vspace{-1em}
\end{table}
\paragraph{CNN-Backbone Mask R-CNN}
We further conduct experiments on Mask R-CNN with CNN backbone. Results in Table~\ref{table: Comparison with Mask R-CNN} compare LSQ baselines trained on 120k/5k real samples from the MS-COCO/Pascal VOC datasets with our method, which uses only 2k/50 synthetic samples.

As presented, we achieve state-of-the-art results on both VOC and MS-COCO datasets, surpassing LSQ trained with full real data at 8-bit width. Specifically, on the smaller VOC dataset, our method surpasses LSQ trained with the entire dataset by 0.5\% while using only 1/100 of the training data, and exceeds LSQ trained on a similar dataset size by 2\%. On the larger MS-COCO dataset, our approach outperforms LSQ trained with the full dataset by 0.2\% using only 1/60 of the training data, and surpasses LSQ trained on a comparable dataset size by 2.3\%. These findings highlight the robustness and strong generalization ability of our method across datasets of varying scales.

\paragraph{Transformer-Backbone Mask R-CNN}
\label{subsubsec: Comparison with vits}

\begin{table}[t]
\caption{Comparison with real data QATs on Transformer-based Mask R-CNN on MS-COCO validation set.}
\label{table: Comparison on vit}
\centering
\resizebox{\linewidth}{!}{
        \begin{tabular}{llllcc}
        \toprule
        & & & & \multicolumn{2}{c}{mAP / mAP50} \\
        \cmidrule(r){5-6}
        Method & Real Data & Num Data & Prec. & Swin-T & Swin-S  \\
        \midrule
        Pre-trained & \checkmark & 120k(full) & FP & 46.0/68.1 & 48.5/70.2  \\
        \midrule
         LSQ & \checkmark & 120k(full) & \multirow{3}{*}{W8A8} & 45.9/68.0 & 48.1/69.7  \\
         LSQ & \checkmark & 2k &  & 44.4/65.9 & 47.0/68.6 \\
         \rowcolor{teal!10} \textbf{Ours} & \texttimes & 2k & & 45.1/66.7 & 47.1/68.8\\
         \midrule
         LSQ & \checkmark & 120k(full) & \multirow{3}{*}{W6A6} & 44.7/66.8 & 47.1/68.8 \\
         LSQ & \checkmark & 2k & & 41.2/62.9 & 44.4/65.9\\
         \rowcolor{teal!10} \textbf{Ours} & \texttimes & 2k & & 42.0/63.0 & 45.1/65.8\\
        \midrule
         LSQ & \checkmark & 120k(full) & \multirow{3}{*}{W4A8} & 45.5/64.7 & 47.8/69.4 \\
         LSQ & \checkmark & 2k &  & 43.3/65.2 &  45.9/67.3\\
         \rowcolor{teal!10} \textbf{Ours} & \texttimes & 2k &  & 43.0/64.2 & 46.2/67.1 \\
         \bottomrule
        \end{tabular}
}

\end{table}
\vspace{-1em}

In this section, we validate the proposed method on Transformer-based object detection networks. Specifically, we use Swin-T/S~\citep{liu2021swin} as the backbone model, combined with the Mask R-CNN prediction head to form our model. The results on the MS-COCO dataset are presented in Table~\ref{table: Comparison on vit}.

Our method can be seamlessly extended to Transformer-based object detection networks. Compared to LSQ trained with the same amount of data, our approach consistently outperforms it by 0.3\%–0.8\% across different bit-widths. While our method shows a slight performance drop of 0.8\%–2.7\% compared to LSQ trained on the full dataset, it significantly improves QAT efficiency by requiring substantially less data.

\subsection{Comparison with Task-Agnostic Methods}
Previous zero-shot quantization works in object detection primarily employ task-agnostic training methods for both image generation and quantization-aware training~\citep{choi2024mimiq, ramachandran2024clamp, li2023psaq}. We are the first to introduce task-specific loss at both stages. 

Through extensive experiments, as shown in the Table~\ref{table: Comparison on task-agnostic}, we demonstrate that incorporating task-specific loss  improves detection performance across various models, including YOLO11-s/m and Swin-T/S, as well as across different quantization levels, such as 6-bit and 8-bit precision.

This improvement is due to two key factors. First, task-specific training loss enriches the image generation process with detailed information, such as bounding box categories, size and coordinates, resulting in a distribution that more closely resembles real images. Second, during quantization-aware training, it enables the model to learn directly from labels, enhancing its ability to extract meaningful information from images.

\begin{table}[t]
\caption{Comparison with Task-Agnostic Methods on MS-COCO validation set. $L_{detect}$ represents the task-specific training loss. All methods use 2k synthetic images for QAT.}
\label{table: Comparison on task-agnostic}
\centering
\resizebox{1\linewidth}{!}{
    \begin{tabular}{llcccc} % Ensure correct number of columns
    \toprule
    ~ & ~ &  \multicolumn{4}{c}{mAP / mAP50} \\
    \cmidrule(r){3-6}
    Method & Prec. & YOLO11-s & YOLO11-m & Swin-T & Swin-S  \\
    \midrule
    Pre-trained & FP & 47.0/65.0 & 51.5/70.0 & 46.0/68.1 & 48.5/70.2  \\
    \midrule
    \rowcolor{teal!10} \textbf{Ours} & \multirow{2}{*}{W8A8} & \textbf{45.6/62.3} & \textbf{50.0/66.5} & \textbf{45.1/66.7} & \textbf{47.1/68.8} \\ 
    \quad w/o $L_{detect}$ & & 43.6/60.2 & 49.8/66.7 & 44.6/66.8 & 45.1/67.1  \\
    \midrule
    \rowcolor{teal!10} \textbf{Ours} & \multirow{2}{*}{W4A8} & \textbf{42.6/58.9} & \textbf{47.7/64.1} & \textbf{43.0/64.2} & \textbf{46.2/67.1} \\ 
    \quad w/o $L_{detect}$ & & 39.7/56.0 & 47.3/64.2 & 43.0/64.8 & 43.6/65.4  \\
    \midrule
    \rowcolor{teal!10} \textbf{Ours} & \multirow{2}{*}{W6A6} & \textbf{43.0/59.3} & \textbf{47.1/63.2} & \textbf{42.0/63.0} & \textbf{45.1/65.8} \\ 
    \quad w/o $L_{detect}$ & & 40.7/57.0 & 46.6/63.2 & 40.4/61.7 & 42.8/64.6  \\
    \bottomrule
    \end{tabular}
}
\end{table}

\subsection{Comparison with Data Free Methods} \label{subsubsec: Data free Sampling}
\begin{table}[t]
\caption{Comparison with Data free Methods on MS-COCO validation set. All methods use 2k synthetic images for W6A6 QAT on YOLOv5-s. Info. denotes detailed information about labels including bouding box categories and coordinates. Distri. represents quantity distribution information about the labels per image. "-" indicates that the network diverges.}
\label{table: Sample effectiveness}
\centering
\resizebox{1\linewidth}{!}{
    \begin{tabular}{lllcc} % Ensure correct number of columns
    \toprule
    Method & Info. & Distri. & mAP & mAP50  \\
    \midrule
    Real Label & \checkmark & \checkmark & 32.7 & 51.4 \\
    \midrule
    \textbf{Gaussian} noise & \texttimes & \texttimes & - & - \\ 
    \textbf{Tile}(Out-of-distri.) & \texttimes & \texttimes & 23.9 & 39.0 \\
    \textbf{Tile}(In-distri.) & \texttimes & \checkmark & 24.0 & 39.3 \\
    \textbf{MultiSample}(Out-of-distri.) & \texttimes & \texttimes & 28.2 & 46.7 \\
    \textbf{MultiSample}(In-distri.) & \texttimes & \checkmark & 29.7 & 48.0 \\
    \rowcolor{teal!10} \textbf{Ours(Adaptive Label Sampling)} & \texttimes & \texttimes & \textbf{32.0} & \textbf{50.0} \\
    \bottomrule
    \end{tabular}
}
\end{table}

Furthermore, we explore a completely data-free scenario where no information about real images or labels is available and demonstrate the robustness of our adaptive label sampling method. The results are presented in Table~\ref{table: Sample effectiveness}.

First, we establish a weak baseline by using \textbf{Gaussian} noise as targets for QAT training and find that the quantized network fails to converge. This highlights the importance of synthetic data quality for QAT. For other proxy datasets, we primarily consider two types: \textit{in-distribution} and \textit{out-of-distribution}. \textit{In-distribution} datasets assume that the number of bounding boxes per image is known, making the generated images statistically closer to real ones. In contrast, \textit{out-of-distribution} datasets assume no prior knowledge about the original labels.

For baseline methods, we include \textbf{Tile}, which divides the image into uniform grids and randomly generates a unique bounding box in each grid, including its category and coordinate information, and \textbf{MultiSample}, which randomly samples multiple labels for each image. As shown in Table~\ref{table: Sample effectiveness}, QAT using images generated by our adaptive label sampling method surpasses the best \textit{in-distribution} proxy dataset by 2.3\% at 6-bit precision. Furthermore, when comparing our sampling method with images generated using real labels, we observe only a 0.7\% performance gap. This demonstrates that our adaptive label sampling method can effectively extract real label information from the network, thus enabling the generation of high-quality synthetic data. Additional results across different precision levels are also provided in Appendix~\ref{subsec: full comparison with data free}.

\section{Conclusions}\label{label:conclusions}
% \vspace{-0.2cm}

For the first time, we revisit the current task-agnostic zero-shot quantization (ZSQ) methods for object detection tasks and identify the inherent limitations in their performance due to their task-agnostic nature. we propose a novel zero-shot quantization framework specifically tailored for object detection. The proposed framework consists of two key components: a novel task-specific synthesis process for generating the calibration set and a task-specific distillation process. The task-specific synthesis process leverages a bounding box and category sampling strategy to extract more relevant information from the original model, while the task-specific distillation process utilizes this information to fine-tune the quantized model, thereby significantly enhancing its performance.
Extensive experiments demonstrate that our proposed method is both efficient and accurate, achieving performance comparable to traditional quantization-aware training (QAT) methods, such as Learned Step Size Quantization (LSQ), which rely on full real data, while significantly outperforming task-agnostic counterparts. This empowers zero-shot quantization with immense practical significance for object detection tasks.

% \clearpage
{
    \small
    \bibliographystyle{ieeenat_fullname}
    \bibliography{main}
}

\clearpage
\appendix
\section{Related Works}\label{sec: related}

This section provides a brief overview of the studies relevant to our work, focusing on data-driven quantization and zero-shot quantization.
\paragraph{Data-driven Quantization}
Post-training quantization (PTQ) and quantization-aware training (QAT)  \citep{white2018,white2021} are the most commonly employed quantization methods. PTQ methods typically utilize a small calibration set, often a subset of the training data, to optimize or fine-tune quantized networks \citep{finkelstein2023ptq,frantar2022optimal}. For instance, AdaRound \citep{nagel2020adaround} introduced a layer-wise adaptive rounding strategy, challenging the quantizers of rounding to the nearest value. Additionally, BRECQ \citep{li2021brecq} implemented block-wise and stage-wise reconstruction techniques, striking a balance between layer-wise and network-wise approaches. QDrop \citep{wei2022qdrop} innovatively proposed randomly dropping activation quantization during block construction to achieve more uniformly optimized weights. Despite their simplicity and minimal data requirements, PTQ methods often face challenges related to local optima due to the limited calibration set available for fine-tuning.
On the other hand, most QAT approaches leverage the entire training dataset to quantize networks during the training process \citep{jung2019learning}.
PACT \citep{choi2018pact} introduced a parameterized clipping activation technique to optimize the activation clipping parameter dynamically during training, thereby determining the appropriate quantization scale. LSQ \citep{esser2019learned} proposed estimating the loss gradient of the quantizer's step size and learning the scale parameters alongside other network parameters. LSQ+ \citep{bhalgat2020lsq+}, an extension of the LSQ method, introduced a versatile asymmetric quantization scheme with trainable scale and offset parameters capable of adapting to negative activations.
Both QAT and PTQ methods rely on training data for quantization, rendering them impractical when faced with privacy or confidentiality constraints on the training data.

\paragraph{Zero-shot Quantization}
Zero-Shot Quantization (ZSQ) is a valuable approach that eliminates access to real training data during the quantization process. Presently, most ZSQ research is confined to classification tasks. Data-free quantization (DFQ) represents a subset of ZSQ methods that enable quantization without relying on any data. For instance, DFQ \citep{nagel2019dfq} introduced a scale-equivariance property of activation functions to normalize the weight ranges across the network. SQuant \citep{guo2022squant} developed an efficient data-free quantization algorithm that does not involve back-propagation, utilizing diagonal Hessian approximation. However, due to the absence of data, DFQ methods may not be suitable for low-bit-width configurations. For example, in the case of 4-bit MobileNet-V1 on ImageNet, SQuant achieved only 10.32\% top-1 accuracy.
Another branch of ZSQ methods leverages synthetic data \citep{chen2023adeq,li2023hard} generated by the full-precision network. GDFQ \citep{xu2020gdfq} introduced a knowledge-matching generator to synthesize label-oriented data using cross-entropy loss and batch normalization statistics (BNS) alignment. TexQ \citep{chen2024texq} emphasized the detailed texture feature distribution in real samples and devised texture calibration for data generation. Recently, the latest works extended ZSQ to more downstream tasks including object detection. PSAQ-ViT V2 \cite{li2023psaq} introduce an adaptive teacher–student strategy to the patch similarity metric in PSAQ-ViT and generate task-agnostic images to fine-tune the quantized model with knowledge distillation for segmentation and object detection. Similarly, MimiQ \cite{choi2024mimiq} proposed inter-shead attention similarity and apply head-wise structural attention distillation to align the attention maps of the quantized network to those of the full-precision teacher across downstream tasks. CLAMP-ViT \cite{ramachandran2024clamp} employed a two-stage approach, cyclically adapting between data generation and model quantization for classification and object detection tasks. However, the synthetic images they use in downstream tasks are task-agnostic, without containing the specific information required for the corresponding task. Our work validates task-specific image lifting performance of quantized model, yielding state-of-the-art results in ZSQ for object detection.

\section{Implementation Details}\label{sec: Implementation Details}
We report mAP and mAP50 on the validation sets of MS-COCO 2017 \citep{lin2014microsoft} and Pascal VOC~\citep{everingham2010pascal} for the object detection task. Our replication experiments not only contain one-stage YOLO networks such as classic YOLOv5 and recent YOLO11 networks, but also include two-stage Mask R-CNN networks with both ResNet and Transformer-based backbones. All experiments are conducted using a pre-trained model as the teacher. The YOLOv5/YOLO11 experiments are executed on two NVIDIA GeForce RTX 4090 GPUs, while the Mask R-CNN/ViT experiments are conducted on 4 and 8 GPUs, respectively.

\subsection{Adaptive Label Sampling } \label{subsec: adaptive label sampling}
\begin{figure*}[t]
    \includegraphics[width=\linewidth]{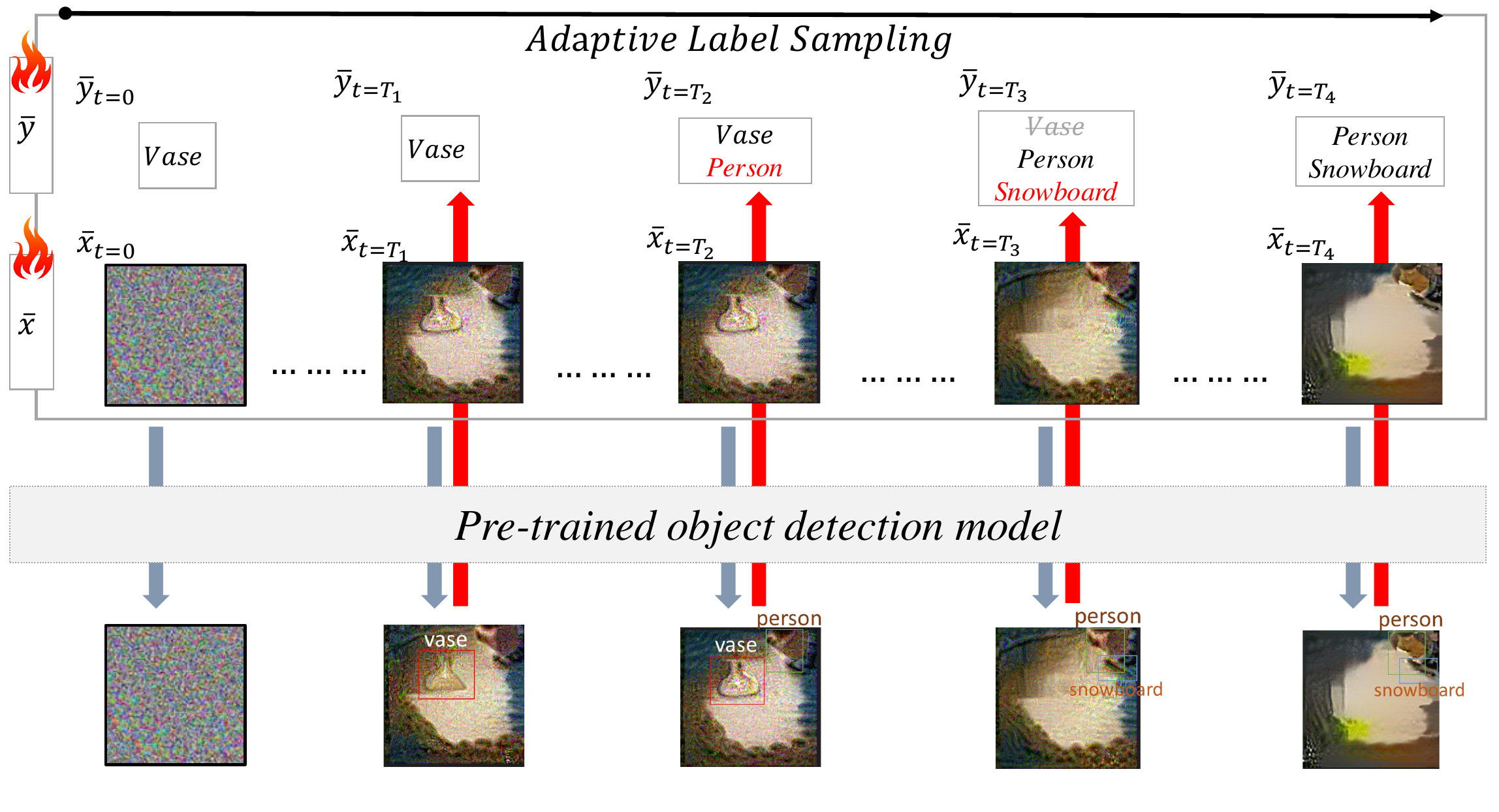}
    \caption{An overview of \textbf{Adaptive Label Sampling} process. We randomly initialize the label $\overline{y}$ and initialize the input image $\overline{x}$ using Gaussian noise. For Every fixed interval, we use a pre-trained object detection model to re-detect objects in $\overline{x}$ and update the target $\overline{y}$. In the subsequent iterations, $\overline{x}$ is optimized toward the updated target $\overline{y}$. We observe that, over iterations, $\overline{x}$ and $\overline{y}$ become increasingly aligned with each other. }
    \label{fig: relabel framework}
    % \vspace{-0.5cm}
\end{figure*}

While theoretically merging labels and image updates into a single stage seems feasible, our experiments in Section~\ref{subsec: adaptive sampling stage} indicate that a continuously evolving target can negatively impact the quality of the final generated images. To address this issue, we first conduct a rapid adaptive label sampling at a low resolution (160) and then use the fixed labels to generate high-resolution images (640). We also provide details of the initial label random generation in adaptive label sampling in Table~\ref{table: One box sampling}, and an outline of the overall algorithm in Algorithm~\ref{alg:adaptive sampling}. Fig.~\ref{fig: relabel framework} provides a visual representation of the algorithm's process.

\begin{table}[t]
\centering
    \vspace{-0.5em}
    % \vspace{-2em}
    \caption{Bounding box sampling details: we start by sampling one object \textbf{Y} for each image, where \textbf{C} represents the number of categories. We assume that the relative width and height of the image are both 1. $W_{\text{min}}$ and $H_{\text{min}}$ are set to 0.2, while $W_{\text{max}}$ and $H_{\text{max}}$ are set to 0.8. $\mathcal{U}$ denotes uniform distribution.}
    \label{table: One box sampling}
    % \resizebox{0.68\linewidth}{!}{
    \begin{tabular}{lll}
    \toprule
    Variable & Sampling Distribution & Description \\
    \midrule
    \textbf{Y}{[}i,0{]} & - & Batch index \\
    \textbf{Y}{[}i,1{]} & $\mathcal{U}(0,C)$ & Category \\
    \textbf{Y}{[}i,2{]} & $\mathcal{U}(W/2, 1-W/2)$ & Bounding box x-center \\
    \textbf{Y}{[}i,3{]} & $\mathcal{U}(H/2, 1-H/2)$ & Bounding box y-center \\
    \textbf{Y}{[}i,4{]} & $\mathcal{U}(W_{\text{min}}, W_{\text{max}})$ & Bounding box width \\
    \textbf{Y}{[}i,5{]} & $\mathcal{U}(H_{\text{min}}, H_{\text{max}})$ & Bounding box height \\
    \bottomrule
    \end{tabular}%}
\end{table}

\begin{algorithm}[H]
\footnotesize
   \caption{Adaptive Label Sampling Algorithm}
   \label{alg:adaptive sampling}
\begin{algorithmic}
   \STATE {\bfseries Input:} current stage image and labels $\{img, tgts\}$, pre-trained detection network $teacher$, filtering threshold: confidence $conf\_thresh$, iou $iou\_thresh$
   \STATE 1. $new\_tgts = teacher(img).predictions[conf > conf\_thresh]$
   \STATE 2. $ious = \operatorname{IOU}(new\_tgts, tgts)$
   \STATE \# Add labels that do not overlap with the existing labels.
   \STATE 3. $add\_tgts = new\_tgts[(\operatorname{max}(ious, dim=1) < iou\_thresh)]$
   \STATE \# Remove labels from the existing list that are not detected by teacher.
   \STATE 4. $minus\_tgts = (\operatorname{max}(ious, dim=0) < iou\_thresh).bool()$
   \STATE 5. $tgts = tgts[\sim minus\_tgts]$
   \STATE 6. $tgts = \operatorname{cat}([tgts, add\_tgts], dim=0)$
\end{algorithmic}
\end{algorithm}

\subsection{Calibration Set Generation} 
 We apply Eq.~\ref{eqn:main loss} and set the optimal trade-off parameters for \{$\alpha_{detect}$,  $\alpha_{BN}$, $\alpha_{TV}$, $\alpha_{l_2}$\} as \{$0.5, 0.01, 0, 5\text{e-}4$\} for the YOLOv5 series model, \{$1\text{e-}3, 1\text{e-}3, 0, 5\text{e-}5$\} for the YOLO11 series model, \{$5.0, 2\text{e-}3, 0, 1.5\text{e-}5$\} for CNN-backbone Mask R-CNN model and \{10.0, 1.0, 0, 1\text{e-}3\} for Transformer-backbone Mask R-CNN model. We generate $X_{inv}$ by optimizing the cost function for 2500/3000/8000/4000 iterations for YOLOv5 series model, YOLO11 series model, CNN-backbone Mask R-CNN model and Transformer-backbone Mask R-CNN model respectively. We use Adam as the optimizer with an initial learning rate of 1e-2, adjusted by cosine annealing~\citep{loshchilov2016sgdr}. We also use cutout~\citep{devries2017improved} as a data augmentation method to enhance the diversity of the synthetic calibration set.

\subsection{Quantization Aware Training}
Since the original $LSQ$ is only evaluated in classification tasks on ImageNet, we extended it to object detection tasks. For each of our networks, $LSQ$ is attached to all internal layers except the first and last layers following ~\citep{esser2019learned}. Our training data are from the synthesized calibration set aforementioned. During $QAT$, we use per-tensor symmetric quantization for both activations and weights and learn the quantization scaling/bias factor via back-propagation with the $Adam$ optimizer. The learning rate for YOLOv5, YOLO11, CNN-backbone Mask R-CNN and Transformer-backbone Mask R-CNN are 1e-4, 1e-5, 1e-4 and 1e-6 respectively. Other experimental hyper-parameters follow official implementations. We use Eq.~\ref{eqn:QAT loss} as our loss function, with optimized hyper-parameters for \{$\beta_{detect}$, $\beta_{KL}$, $\beta_{feat}$\} being \{$0.04, 0.1, 1.0$\} for the YOLOv5 series model, \{ 0.01, 0.1, 1.0 \} for the YOLO11 series model, \{$0.04, 0.2, 0.1$ \} for CNN-backbone Mask R-CNN model, and \{$1.0, 1.0, 1.0$\} for Transformer-backbone Mask R-CNN model. After training, we report mAP/mAP50 as our results.

% Rest experimental hyper-parameters follow official YOLOv5\footnote{\href{https://github.com/ultralytics/yolov5/tree/master}

\section{Ablation Study}\label{sec: Ablation study}
% \changhao{Check use 4-bit/5-bit, which is more appropriate}

% \changhao{should we add algorithm description and one box sampling strategy and more visualization of different neural networks? Should we keep comparing with Nvidia pics in the following pictures?}

\subsection{Adaptive Label Sampling Stage} \label{subsec: adaptive sampling stage}
% \changhao{Check use 4-bit/5-bit, which is more appropriate}
We conduct ablations on the impact of the sampling stage number for the YOLOv5-s model, results are shown in Table~\ref{table: Ablations on Sampling Stage}. Overall, the two-stage sampling strategy outperforms the one-stage strategy, which we attribute to the continuous variation of targets causing fluctuation in the regression targets of the image, thus hindering stable convergence. It also matches the performance of the three-stage approach. Ultimately, we opt for the two-stage strategy to strike a balance between performance and cost.

% \changhao{Hopefully: as shown in Table~. Next sentence: based on whether three-stages are significantly better than two-stages result}

\begin{table}[h]
\caption{Ablations on Adaptive Label Sampling stages number. One stage: update images and labels simultaneously in one process. Two stages: Sample out labels first, then synthesize images with fixed labels. Three stages: Generate images with one random label first, then sample out labels with fixed images, and finally synthesize images with fixed labels}
    \label{table: Ablations on Sampling Stage}
    \centering
        \begin{tabular}{llll}
        \toprule
        Stages Num & Precision & mAP & mAP50 \\
        \midrule
         & W6A6 & 30.6 & 48.8 \\
        One & W5A5 & 25.2 & 41.1 \\
         & W4A4 & \textbf{16.0} & \textbf{27.9} \\
         \midrule
         & W6A6 & \textbf{32.1} & \textbf{50.1} \\
        Two & W5A5 & \textbf{26.3} & \textbf{42.3} \\
         & W4A4 & 15.8 & 28.1 \\
         \midrule
         & W6A6 & 31.7 & 49.3 \\
        Three & W5A5 & 26.1 & 42.5 \\
         & W4A4 & 15.7 & 27.8\\
         \bottomrule
        \end{tabular}
\end{table}

\subsection{Calibration Set Size} After hyper-parameters are fixed, the calibration set size $S$ is searched for its optimal trade-off between computation cost and effectiveness with grid search by quantizing YOLOv5-s to 4-8 bits, as displayed in Table~\ref{table: Ablations on Calibration Set Size}. When $S$ reaches 2k, the performance of the quantized network approaches convergence. Further increasing the size will lead to increased data generation time and computational costs. To avoid complex searches, the same $S$ is used for all experiments. While this may not be optimal for all networks, it is sufficient to demonstrate the superiority of our approach.

\begin{table*}[t]
    \caption{A detailed analysis of calibration set size $S$ across different bit widths using YOLOv5-s on MS-COCO validation set.}
    \label{table: Ablations on Calibration Set Size}
    \centering
    % \begin{footnotesize}
    % \begin{sc}
    % \resizebox{0.6\linewidth}{!}{
        \begin{tabular}{llllllll}
        \toprule
        ~ & ~ & ~ & \multicolumn{5}{c}{mAP} \\
        \cmidrule(r){4-8}
        Method & Real Data & $S$ & W4A4 & W5A5 & W6A6 & W7A7 & W8A8 \\
        \midrule
        LSQ & \checkmark & 120k (Full) & 23.3 & 26.9 & 31.5 & 33.4 & 35.7 \\
        \midrule
         & \texttimes & 5k & 19.1 & \textbf{28.0} & 32.6 & 34.9 & 35.7 \\
         & \texttimes & 4k & 18.9 & 27.9 & \textbf{32.8} & 34.7 & 35.8 \\
        Ours & \texttimes & 3k & \textbf{19.2} & 27.9 & 32.7 & \textbf{35.0} & \textbf{36.0} \\
         & \texttimes & 2k & 19.0 & 27.4 & 32.7 & 34.7 & 35.4 \\
         & \texttimes & 1k & 18.3 & 27.8 & 32.6 & 34.8 & 35.6 \\
         \bottomrule
        \end{tabular}
        % }
    %     \end{sc}
    % \end{footnotesize}
    % \vspace{-0.5cm}
    \end{table*}

\subsection{Modules} Ablation on key modules of the QAT stage including $\mathcal{L}_{KD}$ (Kullback-Leibler Divergence), $\mathcal{L}_{detect}$ , and $\mathcal{L}_{feat}$ is conducted. As presented in Table~\ref{table: Ablations on modules}, dropping one or two of them results in a mAP loss. The largest mAP loss~(7.2\%) occurs when both $\mathcal{L}_{KD}$ and $\mathcal{L}_{feat}$ are removed, indicating their cooperative relationship: $\mathcal{L}_{feat}$ constrains features of network layers, facilitating $\mathcal{L}_{KD}$ to align the network's predictions with the targets.

\begin{table}[h]
    \caption{Ablations on modules. We use 2k calibration set and report mAP/mAP50 of 4-bit YOLOv5-s on MS-COCO validation set.}
    \label{table: Ablations on modules}
    \centering
    % \begin{footnotesize}
    % \begin{sc}
    % \resizebox{0.5\linewidth}{!}{
        \begin{tabular}{lllll}
            \toprule
            $\mathcal{L}_{feat}$ & $\mathcal{L}_{KD}$ & $\mathcal{L}_{detect}$ & mAP & mAP50 \\
            \midrule
            \checkmark & \checkmark & \checkmark & \textbf{19.0} & \textbf{33.4} \\
             & \checkmark & \checkmark & 16.8 & 30.1 \\
             &  & \checkmark & 11.8 & 21.5 \\
             \bottomrule
            \end{tabular}
            % }
    %     \end{sc}
    % \end{footnotesize}
    % \vspace{-0.5cm}
    \end{table}

\subsection{Full Comparison with Data-Free Methods} \label{subsec: full comparison with data free}
In this section, we present a comparison of our adaptive label sampling method with other baseline methods under the data-free scenario across multiple precision levels. From the results in Table~\ref{table: Full Data Free}, we observe that despite being restricted from accessing specific real label information or distribution details, our method consistently outperforms other data-free approaches. Moreover, it achieves results comparable to those obtained using real labels across different precision levels.

\begin{table}[ht]
\caption{Comparison with Data free Methods on MS-COCO validation set across multiple precision levels. All methods utilize 2k synthetic images for QAT on YOLOv5-s.}
\label{table: Full Data Free}
\centering
\resizebox{1\linewidth}{!}{
    \begin{tabular}{llllcc} % Ensure correct number of columns
    \toprule
    Prec. & Method & Real label & Data distri. & mAP & mAP50  \\
    \midrule
    \multirow{7}{*}{W5A5} & Real Label & \checkmark & \checkmark & 28.0 & 45.8 \\
    \cmidrule(r){2-6}
    & Gaussian noise & \texttimes & \texttimes & - & - \\ 
    & Tile(Out-of-distri.) & \texttimes & \texttimes & 16.1 & 27.9 \\
    & Tile(In-distri.) & \texttimes & \checkmark & 17.7 & 31.0 \\
    & MultiSample(Out-of-distri.) & \texttimes & \texttimes & 21.9 & 37.3 \\
    & MultiSample(In-distri.) & \texttimes & \checkmark & 22.5 & 37.4 \\
    &  \textbf{Ours(Adaptive Label Sampling)} & \texttimes & \texttimes & \textbf{26.1} & \textbf{42.3} \\
    \midrule
    \multirow{7}{*}{W4A4} & Real Label & \checkmark & \checkmark & 19.0 & 33.4 \\
    \cmidrule(r){2-6}
    & Gaussian noise & \texttimes & \texttimes & - & - \\ 
    & Tile(Out-of-distri.) & \texttimes & \texttimes & 5.4 & 11.1 \\
    & Tile(In-distri.) & \texttimes & \checkmark & 6.8 & 13.4 \\
    & MultiSample(Out-of-distri.) & \texttimes & \texttimes & 11.9 & 22.3 \\
    & MultiSample(In-distri.) & \texttimes & \checkmark & 13.1 & 23.3 \\
    &  \textbf{Ours(Adaptive Label Sampling)} & \texttimes & \texttimes & \textbf{15.0} & \textbf{27.0} \\
    \bottomrule
    \end{tabular}
}
\end{table}

\subsection{Comparison with YOLOv5 at Lower Precision} \label{full comparison with yolov5}
In Table~\ref{table: Comparison on YOLO lower bit}, we provide additional performance results of the YOLOv5 series models on the MS-COCO dataset at lower precision levels. Notably, even at ultra-low 4-bit precision, our method still outperforms LSQ trained with full data in most cases. For example, on YOLOv5-l, our approach surpasses the best baseline by 1.7\%, with the gap further increasing to 6.3\% at 5-bit precision.

\begin{table}[ht]
% \vspace{-4em}
\caption{Comparison with real data QATs on YOLOv5 on MS-COCO validation set.}
\label{table: Comparison on YOLO lower bit}
\centering
% \begin{sc}
\resizebox{1\linewidth}{!}{
        \begin{tabular}{llllccc}
        \toprule
        ~ & ~ & ~ & ~ & \multicolumn{3}{c}{mAP / mAP50} \\
        \cmidrule(r){5-7}
        Method & Real Data & Num Data & Prec. & YOLOv5-s & YOLOv5-m & YOLOv5-l \\
        \midrule
        Pre-trained & \checkmark & 120k(full) & FP & 37.4/56.8 & 45.4/64.1 & 49.0/67.3\\
        \midrule
         LSQ & \checkmark & 120k(full) & \multirow{5}{*}{W5A5} & 26.9/44.9 & 32.9/50.6 & 35.2/53.0 \\
         LSQ+ & \checkmark & 120k(full) &  & 27.0/44.9 & 33.1/51.0 & 35.2/53.4 \\
         LSQ & \checkmark & 2k &  & 24.7/42.2 & 31.2/49.3 & 35.2/53.1\\
         LSQ+ & \checkmark & 2k &  & 25.0/42.9 & 31.2/49.2 & 34.8/52.7 \\
         \rowcolor{teal!10} \textbf{Ours} & \texttimes & 2k &  & \textbf{28.0}/\textbf{45.8} & \textbf{37.1}/\textbf{55.7} & \textbf{41.5}/\textbf{59.7} \\
        \midrule
         LSQ & \checkmark & 120k(full) & \multirow{5}{*}{W4A4} & 23.3/40.0 & 27.9/45.4 & 33.1/50.3\\
         LSQ+ & \checkmark & 120k(full) &  & \textbf{23.3}/\textbf{40.2} & 27.7/44.6 & 33.3/50.9 \\
         LSQ & \checkmark & 2k &  & 17.2/32.2 & 25.5/42.3 & 28.9/45.7 \\
         LSQ+ & \checkmark & 2k &  & 17.3/32.1 & 26.1/42.6 & 28.6/45.8 \\
         \rowcolor{teal!10} \textbf{Ours} & \texttimes & 2k &  & 19.0/33.4 & \textbf{29.5}/\textbf{47.1} & \textbf{35.0}/\textbf{52.6} \\
         \bottomrule
        \end{tabular}
}

\end{table}

\subsection{Comparison with Other ZSQ methods}
In this section, we compare our approach with two widely-used zero-shot quantization (ZSQ) methods, Genie~\cite{jeon2023genie} and ZeroQ~\cite{cai2020zeroq}, to highlight the impact of incorporating task-specific information. Notably, both baselines are \textbf{task-agnostic} and therefore \textbf{lack essential knowledge} specific to object detection. As shown in Table~\ref{tab:comparison_with_zsq}, task-specific information plays a critical role in quantization-aware training (QAT). While all methods perform comparably under W4A8 post-training quantization (PTQ), our method achieves a significant \textbf{3.3\% improvement in mAP after QAT}, outperforming all task-agnostic baselines.

\begin{table}[t]
    \caption{\textbf{The validity of task-specific information. (Ours)} mAP/mAP50 of YOLOv11s model on the MS-COCO validation set is reported. Experiments are conducted under the same PTQ or QAT settings, with 512 images from different ZSQ methods.}
    \centering
    \small
    \resizebox{1\linewidth}{!}{
    \begin{tabular}{ccccc}
    \toprule
        Precision & Quantizer setting & Genie~\cite{jeon2023genie} & ZeroQ~\cite{cai2020zeroq} & \textbf{Ours} \\
        \toprule
        \multirow{2}{*}{W8A8}  & PTQ & 45.8/62.3 & 46.0/62.5 & \textbf{46.0}/\textbf{62.7} \\
          &QAT& 39.8/54.6 & 43.9/60.4 & \textbf{45.9}/\textbf{62.2}  \\\midrule
        \multirow{2}{*}{W6A6}  & PTQ & 39.7/54.9 & 40.1/55.5 & \textbf{40.3}/\textbf{55.8} \\
        &QAT & 36.9/51.9 & 39.5/55.4 & \textbf{42.8}/\textbf{59.2} \\\bottomrule
        \multirow{2}{*}{W4A8}  & PTQ & 11.2/18.0 & \textbf{11.3}/\textbf{18.3} & 11.2/18.2 \\
        &QAT & 34.6/49.1 & 37.9/53.9 & \textbf{41.2}/\textbf{57.1} \\\bottomrule
    \end{tabular}}
    \label{tab:comparison_with_zsq}
\end{table}

\section{Sample Efficiency}\label{sec: Sample Efficiency}
\begin{figure}[t]
\vspace{-0.1cm}
    \centering
        \centerline{\includegraphics[width=\columnwidth]{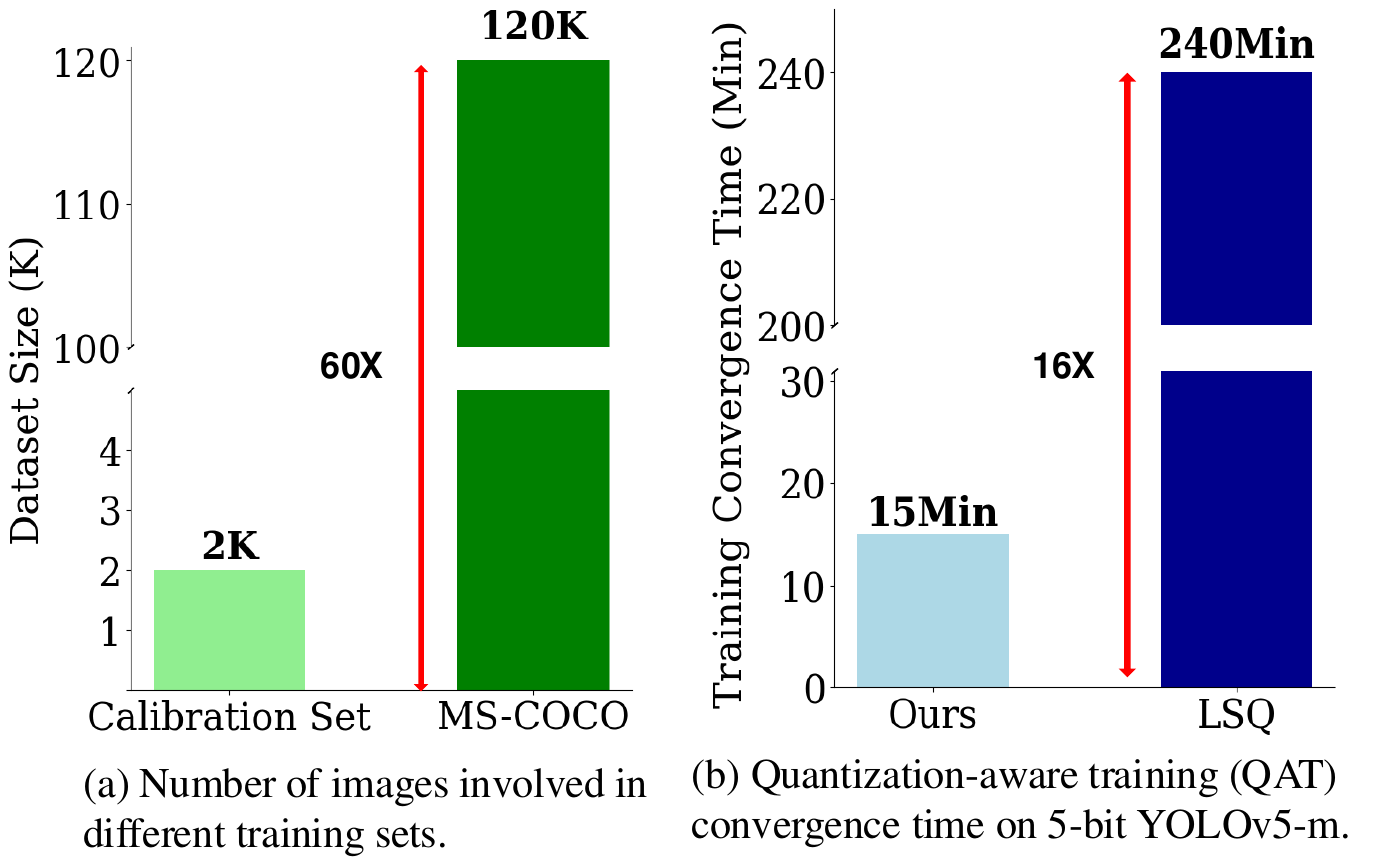}}
    \caption{(a) Our synthetic condensed calibration set is 1/60 the size of the MS-COCO training set. (b) The training convergence speed can be improved by up to 16$\times$ compared to LSQ.}
    % \changhao{how to add caption on subgraphs? Write together first}
    \label{fig: efficiency}
    % \vspace{-0.5cm}
\end{figure}

We also demonstrate that by employing \textbf{Adaptive Label Sampling}, we achieved comparable or even superior results on QAT using a synthetic calibration set that is only \boldmath$1/60$ the size of the original training dataset. Additionally, by integrating self-distillation into the fine-tuning process of the quantized object detection network, we enabled a more efficient knowledge transfer. In the initial stage, utilizing 8 RTX 4090 GPUs for image generation allow us to produce 256 images every 20 minutes, resulting in a total of 160 minutes to generate 2,000 images. It is important to note that the calibration set we generate captures the overall characteristics of the original training set, allowing it to be reused multiple times during the quantization-aware training process. As the number of training iterations increases, our method progressively enhances the training convergence speed, achieving up to \textbf{$16\times$} faster convergence compared to the LSQ method trained on the entire real dataset. The corresponding results are visually illustrated in Fig.~\ref{fig: efficiency}.

\section{Additional Qualitative Results} \label{sec: additional qualitative results}
\paragraph{Visualization for different object detection models}
\begin{figure*}[t]
\vspace{-0.1cm}
    \centering
        \centerline{\includegraphics[width=0.99\linewidth]{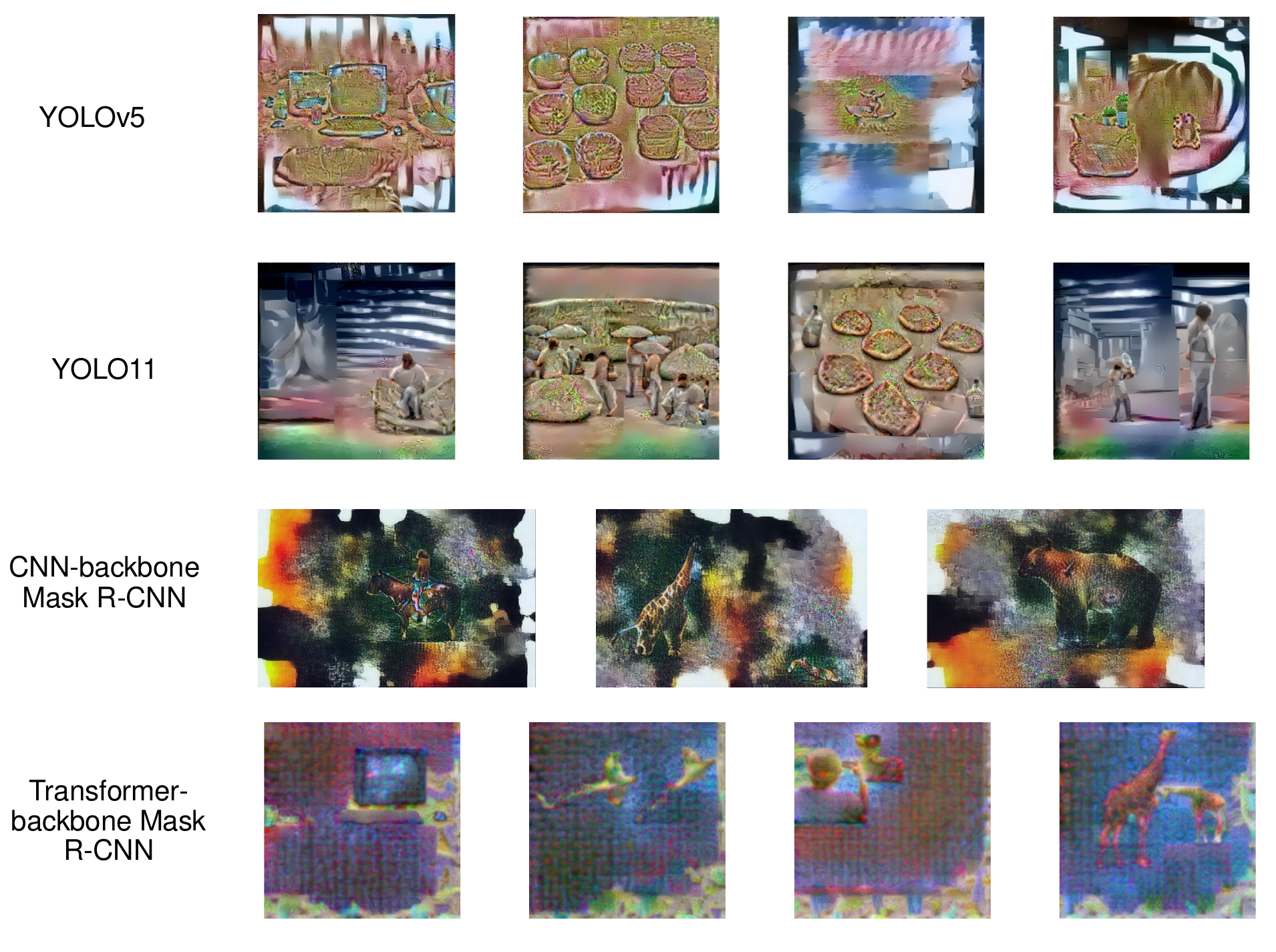}}
    \caption{Visualization of images composed by different architecture-based object recognition networks.}
    \label{fig: img visualization}
    \vspace{-0.5cm}
\end{figure*}

In this section, we present visualizations of images generated by all the models discussed in this paper, including YOLOv5, YOLO11, as well as CNN and Transformer-backbone Mask R-CNN, as shown in Fig.~\ref{fig: img visualization}.
% \changhao{add image visualization}

As observed, despite initializing the images with Gaussian noise and without referencing any real image data, our generated images can still accurately restore the real-world positions and sizes of objects. For instance, in the images generated by YOLOv5, a ballet dancer can be seen gracefully performing. In the images produced by YOLO11, there is a person sitting on a bench waiting for someone, as well as a table with several pizzas on it. In the CNN-backbone Mask R-CNN generated images, a herdsman is riding a horse while a giraffe roams aimlessly. Meanwhile, in the Transformer-backbone Mask R-CNN generated images, a television is displaying a program. By leveraging task-specific image generation, we can create more realistic images across different network architectures without relying on any external support.

\paragraph{Qualitative results for synthetic data} \label{subsec: compare with multi-label}

\begin{figure*}[h]
% \vspace{-0.1cm}
    \centering
        \centerline{\includegraphics[width=0.89\linewidth]{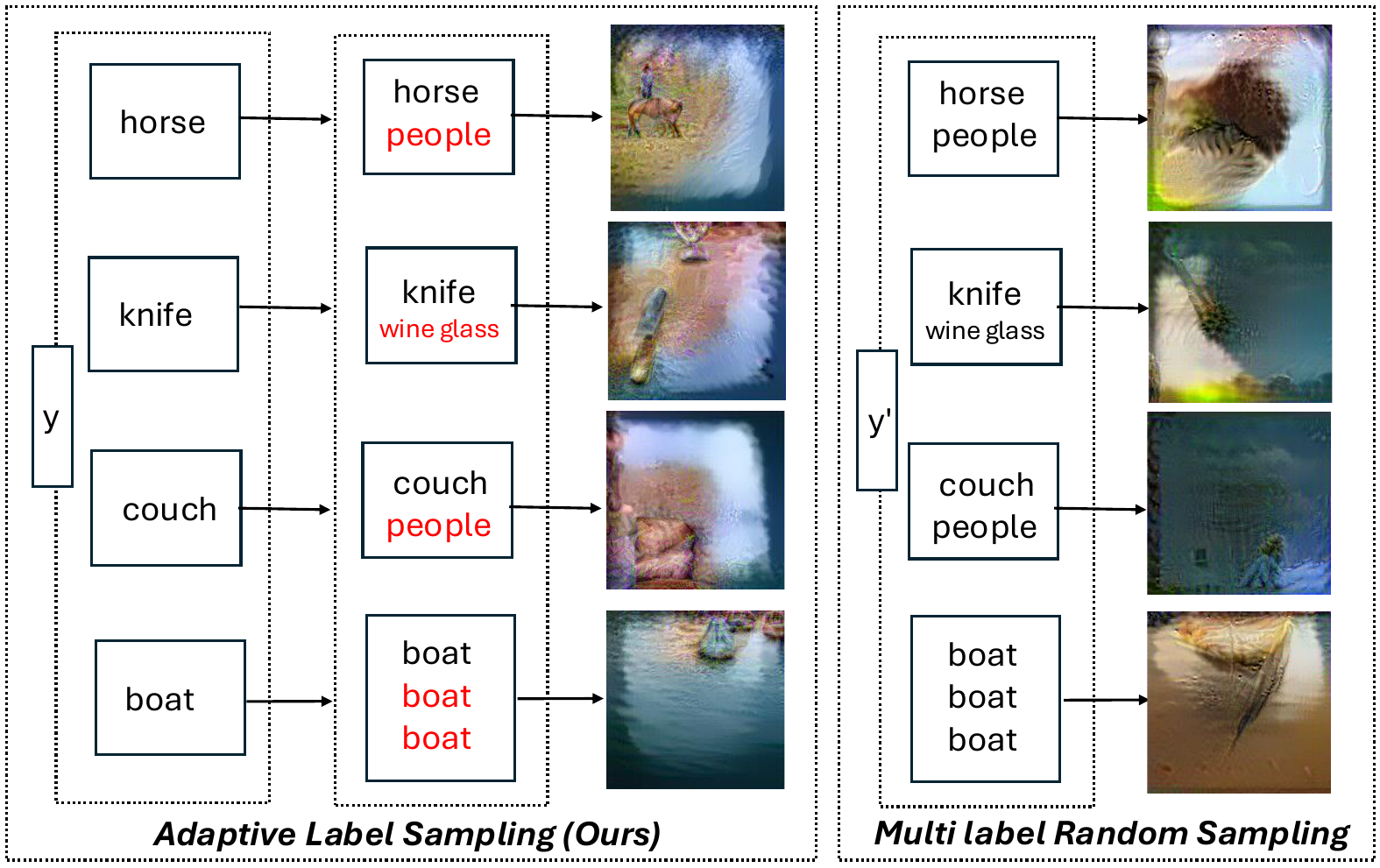}}
    \caption{A comparison of the image quality generated by various sampling methods.}
    \label{fig: comparison with multi-label}
    % \vspace{-0.5cm}
\end{figure*}

In this section, we visualize the advantage of our Adaptive Label Sampling method over random sampling for multiple labels. As shown in Fig.~\ref{fig: comparison with multi-label}, the left side illustrates our Adaptive Label Sampling method, which initially starts with single-label random sampling as presented in Table~\ref{table: One box sampling}. After Adaptive Label Sampling, the model leverages the information stored during pre-training and add objects it considers highly confident, ultimately producing high-quality images. For instance, you can observe a person riding a horse, three boats gently floating on the shimmering water, and someone about to sit and rest next to a couch, among other realistic scenes.

Next, we use the obtained labels to perform multi-label random sampling by generating the corresponding object sizes and locations based on the sampling distribution in Table~\ref{table: One box sampling}. The resulting images are shown on the right side of Fig.~\ref{fig: comparison with multi-label}. In this scenario, the image quality deteriorates significantly, and the visual features fail to accurately reflect the generated objects. Consequently, compared to multi-label sampling, our Adaptive Label Sampling method captures the model's internal information more effectively, producing higher-quality and more coherent images.

\paragraph{Qualitative results for object detection performance}
\begin{figure*}[h]
\vspace{-0.1cm}
    \centering
        \centerline{\includegraphics[width=0.89\linewidth]{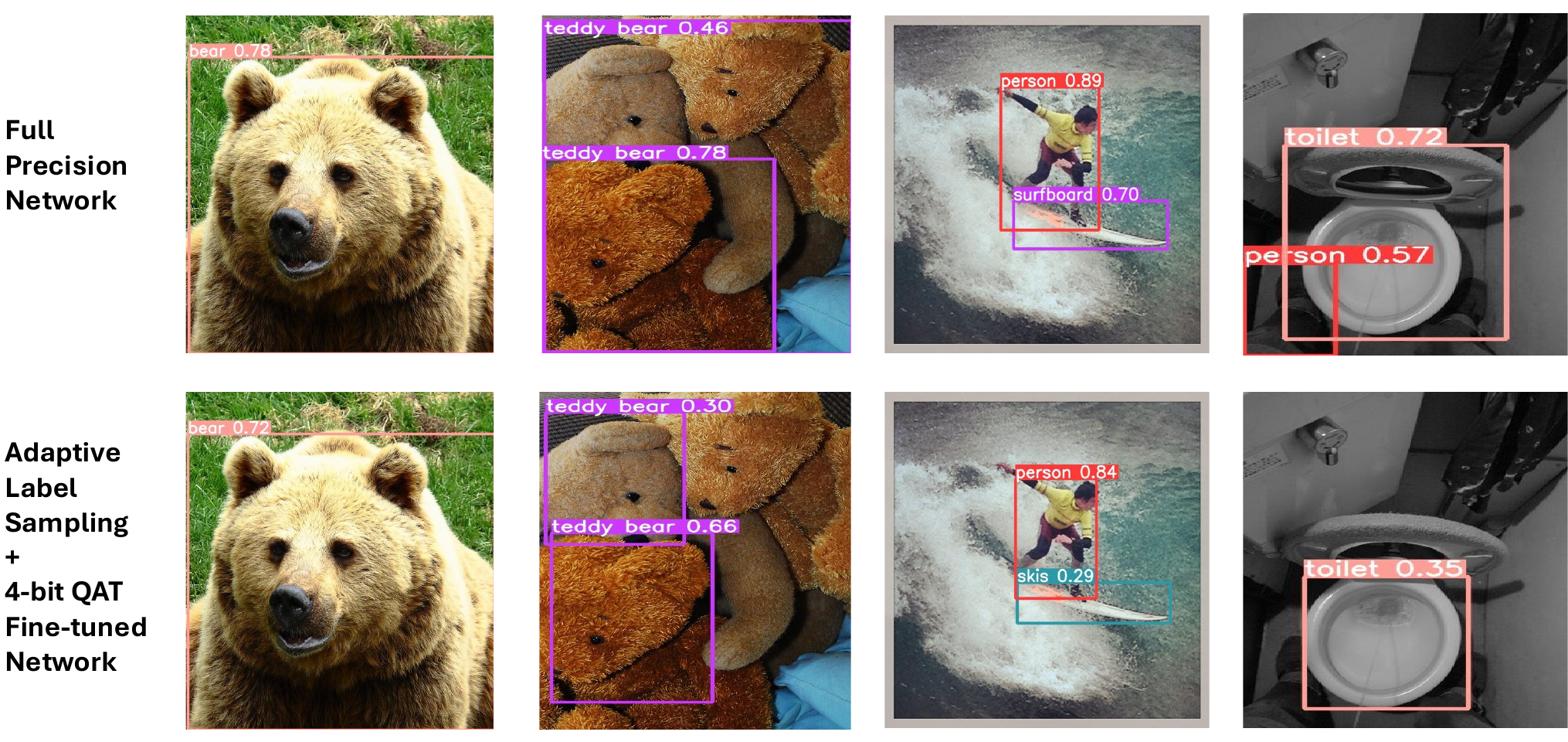}}
    \caption{Qualitative analysis of object detection performance across different neural networks}
    \label{fig: qualitative_network}
    \vspace{-0.5cm}
\end{figure*}

In this section, we present visualizations illustrating the object detection capabilities of various neural networks. Specifically, we randomly selected four images from the MS-COCO validation set and used the detection results of a full-precision YOLOv5-s network as the reference. The visual comparisons in Fig.~\ref{fig: qualitative_network} display the detection results of neural networks trained with our adaptive label sampling method under 4-bit quantization-aware training (QAT).

The visualizations demonstrate that our quantized network can effectively detect objects. For example, when multiple teddy bears are present in an image, it accurately identifies each one. Similarly, when there is only a single object, such as a bear, it correctly recognizes it with confidence levels comparable to those of the full-precision network.

% \ref{subsec: adaptive label sampling}

\end{document}